\PassOptionsToPackage{bookmarks=false}{hyperref}
\documentclass[conference]{IEEEtran}
\IEEEoverridecommandlockouts

\usepackage{cite}
\usepackage{amsmath,amssymb,amsfonts}
\usepackage{graphicx}
\graphicspath{{./images/}}
\usepackage{textcomp}
\usepackage{xcolor}
\def\BibTeX{{\rm B\kern-.05em{\sc i\kern-.025em b}\kern-.08em
    T\kern-.1667em\lower.7ex\hbox{E}\kern-.125emX}}
\usepackage{algorithm}
\usepackage[noend]{algpseudocode}
\usepackage{flushend}
\usepackage{bm}
 \usepackage[caption=false,font=footnotesize]{subfig}
\begin{document}

\title{Adaptive Coordinated Motion Control for Swarm Robotics Based on Brain Storm Optimization \thanks{Correspondance: Yuhui Shi.
This work is partially supported by National Key R\&D Program of China under the Grant No. 2017YFC0804002, National Science Foundation of China under grant number 61761136008, Shenzhen Peacock Plan under Grant No. KQTD2016112514355531, Program for Guangdong Introducing Innovative and Entrepreneurial Teams under grant number 2017ZT07X386, and the Science and Technology Innovation Committee Foundation of Shenzhen under the Grant No. ZDSYS201703031748284 and JCYJ20200109141235597.}  }

\author{\IEEEauthorblockN{Jian Yang}
\IEEEauthorblockA{\textit{Department of Computer Science and Engineering} \\
\textit{Southern University of Science and Technology}\\
Shenzhen, China \\
yangj33@sustech.edu.cn}
\and
\IEEEauthorblockN{Yuhui Shi}
\IEEEauthorblockA{\textit{Department of Computer Science and Engineering} \\
\textit{Southern University of Science and Technology}\\
Shenzhen, China \\
shiyh@sustech.edu.cn}

}

\maketitle

\begin{abstract}
  Coordinated motion control in swarm robotics aims to ensure the coherence of members in space, i.e., the robots in a swarm perform coordinated movements to maintain spatial structures. This problem can be modeled as a tracking control problem, in which individuals in the swarm follow a target position with the consideration of specific relative distance or orientations. To keep the communication cost low, the PID controller can be utilized to achieve the leader-follower tracking control task without the information of leader velocities. However, the controller's parameters need to be optimized to adapt to situations changing, such as the different swarm population, the changing of the target to be followed, and the anti-collision demands, etc. In this letter, we apply a modified Brain Storm Optimization (BSO) algorithm to an incremental PID tracking controller to get the relatively optimal parameters adaptively for leader-follower formation control for swarm robotics. Simulation results show that the proposed method could reach the optimal parameters during robot movements. The flexibility and scalability are also validated, which ensures that the proposed method can adapt to different situations and be a good candidate for coordinated motion control for swarm robotics in more realistic scenarios.
\end{abstract}

\begin{IEEEkeywords}
  Swarm Robotics, Brain Storm Optimization, Formation Control, Leader-follower, Opmimal PID Control
\end{IEEEkeywords}

\section{Introduction}
By taking inspiration from social insects or animals' self-organized behaviors, swarm robotics is a particular method to realize collaborative behaviors of multi-robot systems \cite{trianni2015fundamental}. It aims to achieve robust and scalable collaborative behavior through simple interaction between members or between members and the environment \cite{tan2016Handbook}. This field has been perceived to have the potential in applications of exploration in unknown environments \cite{yang2016formation}, searching particular signal or feature \cite{yang2020exploration,yang2019Extended}, military defense or disaster rescue, etc \cite{fierro2018multi,weinstein2018visual}.  Coordinated motion control in swarm robotics aims to ensure the coherence of members in space, i.e., the robots in a swarm perform coordinated movements to maintain spatial structures \cite{liu2012Coordinated}.

Many works in the literature have designed both loosen and rigid coordinated motion control for robotic swarms. For example, some works dealing with the flocking behaviors, which mimics the self-organized behavior of bird flocks or fish schools, belong to loosen coordinated motion control \cite{stranieri2011self,turgut2008Selforganizeda}. The goal is to let the swarm moving together without collision, without concern about the specific relative positions and orientations among swarm members. Reynolds proposed a representative study using simple rules to achieve flocking: collision avoidance with nearby flockmates, velocity matching with nearby flockmates, and flocking centering with nearby flockmates \cite{reynolds1987Flocks}. From the engineering perspective, to make a group of robots more useful, some other works were presented to deal with the rigid moving formation control problems \cite{brambilla2013Swarm}. Those methods are taking inspiration from both macroscopic and multicellular mechanisms \cite{oh2017Bioinspired}, such as leader-follower structure \cite{liu2019VisionBased}, virtual structure \cite{lewis1997High}, potential field\cite{spears2012Physicomimetics}, etc. Among them, the leader-follower structure has been wildly used \cite{yang2018line}.  It can be modeled as a tracking control problem, in which individuals in the swarm follow a target position with the consideration of specific relative distance or orientations \cite{jiang2018distributed}.

However, some control strategies, such as the traditional leader-follower approach, need the leader's velocity to be exchanged with the followers, which increased the communication costs. Some works dealt with this problem by introducing a PID control law into the leader-follower structure to avoid velocity communication \cite{shen2014Adaptive}, but the PID parameters can not be adjusted adaptively in case of environmental or swarm changes during movements. To compensate for this shortcoming, this paper propose an adaptive online-tuning formation control approach. The proposed method uses the BSO algorithm to refine a set of optimal PID parameters of the tracking controller during movements. The BSO algorithm is a type of swarm intelligence optimization algorithm inspired by human collaborative problem-solving, i.e., the brainstorming process \cite{shi2011Optimization}. Through a series of iterative operations of convergence and divergence, it can find a relatively optimal solution to a specific problem. The BSO has been widely used in many aspects such as data science \cite{yang2020swarm}, electric power systems \cite{mahdad2015security}, optimal control \cite{qiu2014receding}, computer vision applications \cite{yang2020visual}, wireless sensor networks \cite{tuba2018energy}, electromagnetic design problems \cite{aldhafeeri2019brain}, as well as multi-robot systems \cite{yang2020brain}. By applying it to the PID based leader-follower control mechanism, the control parameters will be continuously refined online as the system runs.

The remainder of this paper is as follows: Section 2 reviews the related works. Section 3 states the problems. Section 4 proposes the proposed BSO based coordinated motion control method. Section 5 gives the simulation results with discussions, followed by conclusion in Section 6.

\section{Related Works}

As mentioned, there are mainly two types of coordinated motion control in mobile swarm robotics: flocking and rigid formation motion. Flocking is a procedure observed when a group of birds is foraging or in flight \cite{reynolds1987Flocks}. Rigid moving formation control is a high level of flocking, and it needs not only to keep close together but also to keep some spatial patterning during moving. The rigid formation control approaches include structured approaches and behavioral approaches \cite{oh2017bio}. The structured approaches include leader-follower structure \cite{yang2018VShapeda}, virtual structure \cite{lewis1997High}, and behavioral approaches contain the probability finite state machine \cite{soysal2005probabilistic}, the potential field methods \cite{gazi2005swarm}, as well as the consensus-based approaches \cite{ren2005consensus}.

The leader-follower structure is an excellent method for rigid formation control. By assigning a leader in the swarm, the formation control problem can be modeled as a tracking control problem, where the follower tracks the leader with a specific distance or angle, known as the $l-\varphi$ control \cite{desai1998controlling}. Firstly, Alur et al. presented a leader-follower control scheme, which calculates the control inputs of a swarm of mobile robots with a unicycle model to achieve the formation \cite{alur2001Framework}. However, this method needs the velocity of the leader to be transmitted to the followers. Shen et al. proposed an adaptive PID control strategy without the leader velocity, which determines the inputs under a feedback control law by measuring the errors to the target pose. It has been proven to be stable under the Lyapunov principle \cite{shen2014Adaptive}. Nevertheless, in this work, the turning of the PID parameters is not optimized for different populations or other changes. Another drawback of the leader-follower structure is that when the leader is unavailable, the whole swarm may fail as well. There are some dynamic leader election strategies proposed to tackle this problem \cite{nebro2013Analysis}.

The virtual structure method makes a group of robots behave as if they are mass points embedded in a rigid structure \cite{lewis1997High}. The advantage of this method is that during the formation and movement of the formation, the robots in a group always maintain this relative geometric relationship, which is easy to analyze its convergence and stability. However, the member robot needs to maintain communication and calculation during formation forming and coordinated motion, introducing high complexity of calculation and communication. It is easy to cause excessive throughput and even make the algorithm invalid in a large scale swarm system.

It should be noted that there are also some strategies inspired by the multi-cellular mechanisms \cite{slavkov2018Morphogenesis,sayama2009swarm}, or designed by evolutionary algorithms \cite{roy2016Study,sathiya2019Evolutionary,capi2012Multiple}. This paper focuses on coordinated tracking control for swarm robotics. We will propose an online tuning adaptive control strategy for the leader-follower structure for rigid formation control. We emphasis the following contributions: 1) An online adaptive incremental PID controller without the leader’s velocity is designed for coordinated control for swarm robotics. 2) The parameters of the controller are continuously optimized during the swarm movements. 3) The proposed method can adapt to different maneuver actions and the number of robots in the swarm, which has good flexibility and scalability.

\section{Problem Statement}

\subsection{Assumptions}
First, we assume the swarm will perform actions in a 2-D space. Furthermore, members in the swarm do not have the ability of global positioning. In order to achieve rigid formations, we assign a global leader in the swarm as guidance, but not all of the members can perceive it at any time. The perception and communication are only taking place between the neighbors in a specific range. For the single robot, we adopt the non-holonomic wheeled robot kinematics for single members, i.e., the unicycle model \cite{qu2009cooperative}:
\begin{equation}
  \label{eq:nonholo}
  \left[\begin{matrix} x_i(t+\Delta t)\\ y_i(t+\Delta t) \\ \alpha_i(t+\Delta t) \end{matrix}\right] =
  \left[\begin{matrix} x_i(t) \\ y_i(t) \\ \alpha_i(t) \end{matrix}\right] +
  \left[\begin{matrix} \cos\alpha_i(t) &  0 \\ \sin\alpha_i(t)  & 0 \\  0 & 1 \end{matrix} \right]\left[\begin{matrix} v \\  \omega \end{matrix}\right] 
\end{equation}
where in $(x_i,y_i,\alpha_i)$, $x_i$, $y_i$ the position coordinates of the robot in the reference coordinate system $XOY$, $\alpha$ is its orientation angle, $v$ and $\omega$ indicate the forward and turing speed respectively in each robot's frame $x_ioy_i$. Assume that the robot is equipped with sensors that can detect other member robots' distance and bearing angle, which denote as $l_{ij}$ and $\varphi_{ij}$ respectively for measured distance and bearing angle of robot $j$ in robot $i$'s field of view, then we have:
\begin{equation}
  \label{eq:2}
  (x_{ij},y_{ij})=(l_{ij} \cos \varphi_{ij},l_{ij} \sin \varphi_{ij})
\end{equation}

The coordinated motion problem now can be translated to control a robot reach a target pose that keeps the distance and bearing angle to a reference position. For the flocking motion control, we can define the reference position as the central location of the members in a robot's field of view, without considering the exact distance and bearing angle to this position. The heading is also the average heading angle of all the members heading received. For the rigid formation control, the members are required to follow a specific member in their field of view. For example, the V-shaped moving formation, which imitates some large birds flying formation, have to keep the exact distance and bearing angle to a neighbor. For this goal of coordinated motion control, the swarm leader firstly broadcasts its heading to the surrounding members. If the leader is in someone's field of view, it will determine its role for which side it would follow the leader. And then broadcast its side-role and received leader headings. Other neighbors who can not see the leader, will synchronize its role to the nearest member who already had a role. Based on the synchronized role, the member will follow the closest member in its top-left (for right side member) or top-right (for left side members), which defined as a cascade approach \cite{yang2018VShapeda}.

\subsection{Tracking Control without Velocity Broadcasting}
\begin{figure}[!htb]
  \centering \includegraphics[width=0.4\textwidth]{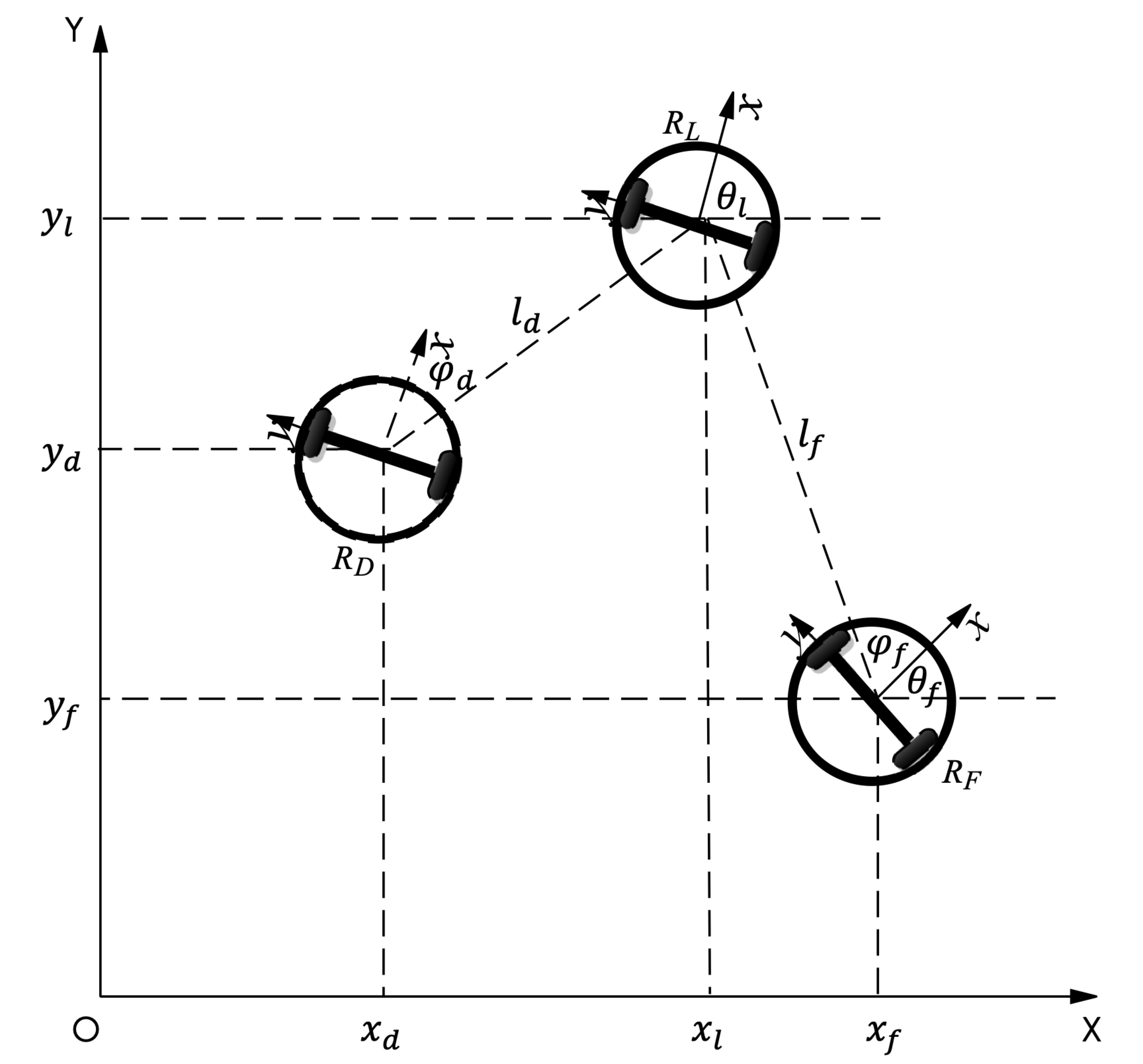}
  \caption{Leader-Follower Structure}
  \label{fig:config}
\end{figure}

As shown in Figure \ref{fig:config}, suppose the follower $R_F$ is
tracking the (virtual) leader $R_L$, and denote the desired position
and pose is $R_D$, then the target position and heading can be
described as:

\begin{equation}
  \label{eq:error1}
  \begin{cases}
    x_d = x_l-l_d\cos (\theta_l+\varphi_d)\\
    y_d = y_l-l_d\sin (\theta_l+\varphi_d)\\
    \theta_d = \theta_l
  \end{cases}
\end{equation}

Then the errors in the reference frame are:
\begin{equation}
  \label{eq:6}
  \begin{cases}
    \Delta X = x_d - x_f \\
    \Delta Y = y_d - y_f \\
    \Delta \Theta = \theta_d - \theta_f
  \end{cases}
\end{equation}

By translating the above errors to the follower's frame ($R_F$), we
have:
\begin{equation}
  \label{eq:error2}
  \left[
    \begin{matrix} e_x \\ e_y \\ e_\theta \end{matrix} \right] =
  \left[\begin{matrix}
      \cos \theta_f & \sin \theta_f & 0 \\
      -\sin \theta_f & \cos \theta_f & 0 \\
      0 & 0 & 1 
    \end{matrix} \right] \left[ \begin{matrix}
      \Delta X \\
      \Delta Y \\
      \Delta \Theta
    \end{matrix}\right]
\end{equation}

In the individual's coordinate system, the forward speed is only
relevant to the error component in the $y$-direction ($e_y$), and the
turning speed is relevant to the error component in the $x$-direction
($e_x$), and the heading error $e_\theta$, then we can use the
following PID controller to achieve the tracking control
\cite{shen2014Adaptive}:

\begin{equation}
  \label{eq:1}
  \left[
    \begin{matrix} v \\ \omega  \end{matrix} \right] =
  \bm{H}(t,\bm{e}) \bm{K}_{pid}(t,\bm{e})
\end{equation}
where:
\begin{equation}
  \label{eq:2}
  \bm{H}(t,\bm{e}) = \left[
    \begin{matrix}
      \bm{E}_x(t) & \bm{0} & \bm{0}  \\
      \bm{0} & \bm{E}_y(t)& \bm{E}_\theta(e)
    \end{matrix}
  \right]
\end{equation}
\begin{equation}
  \label{eq:k1}
  \bm{K}_{pid}(t,\bm{e}) = 
  \left[
    \begin{matrix}
      \bm{K}_x & \bm{K}_y & \bm{K}_\theta
    \end{matrix}
  \right]
\end{equation}
\begin{equation}
  \label{eq:e1}
  \bm{E}_\gamma(t) =\left[ \begin{matrix} e_\gamma(t) & \int e_\gamma(t) &
      \dot e_\gamma(t) \end{matrix}\right]\\
\end{equation}
\begin{equation}
  \label{eq:k2}
  \bm{K}_\gamma(t) =\left[ \begin{matrix} k_{\gamma p}(t) & k_{\gamma i}(t) &
      k_{\gamma d}(t) \end{matrix}\right]
\end{equation}

In equation (\ref{eq:e1}) and (\ref{eq:k2}), we use
$\gamma \in \{x,y,\theta\}$ to represent the parameters relevant to
the components corresponding to $x$, $y$, and $\theta$.
Futhermore, it is unnecessary to remember each one's historical
positions and headings in real applications, so the incremental form
of the PID controller is adopted in this paper, which needs only the
information of the past two time slots. With discrete consideration,
it can be expressed as:
\begin{equation}
  \label{eq:10}
  \left[
    \begin{matrix} \Delta v \\ \Delta \omega  \end{matrix} \right] =
  \Delta\bm{H}(k,\bm{e}) \bm{K}_{pid}(k,\bm{e})
\end{equation}

\begin{equation}
  \label{eq:5}
  \Delta \bm{H}(k,\bm{e}) = \left[
    \begin{matrix}
      \Delta\bm{E}_x(k) & \bm{0} & \bm{0}  \\
      \bm{0} &\Delta \bm{E}_y(k)& \Delta \bm{E}_\theta(k)
    \end{matrix}
  \right]
\end{equation}
where:
\begin{equation}
  \Delta \bm{E}_\gamma(k) =\left[ \begin{matrix}e_\gamma(k)-e_\gamma(k-1) \\ e_\gamma \\ e_\gamma(k)-2e_\gamma(k-1)+e_\gamma(k-2) ] \end{matrix}\right]
\end{equation}

\begin{equation}
  \label{eq:k22}
  \bm{K}_\gamma(k) =\left[ \begin{matrix} k_{\gamma p}(k) & k_{\gamma i}(k) &
      k_{\gamma d}(k) \end{matrix}\right]
\end{equation}
where $\gamma \in \{x,y,\theta\}$.

\subsection{Objective Function}
For equation (\ref{eq:5}), we can use the traditional PID tuning
method to reach an available parameter set, but in our situation, in
order to get good scalability, the population size may change, i.e.,
the swarm is required to work under any number of
members. Furthermore, during the motion, the target position and pose
may dynamically change. We can not use a unique parameter set for all
circumstances. Consequently, the controller parameters $\bm{K}_{pid}$
need to be optimized during moving. Since the objective for the
tracking control is to keep the tracking errors as small as possible,
so the objective function for optimization can be simply written as:
\begin{equation}
  \label{eq:4}
  \begin{aligned}[t]
    & \min_{\bm{K}_{pid}}||\bm{e}(v,w)||, \quad s.t. & v\leq v_{max}, w\leq
    w_{max}
   \end{aligned}
\end{equation}
where $\bm{e}(v,w) = [X_e(v,w)\quad Y_e(v,w) \quad \Theta_e(v,w)]$,
$v_{max}$ and $w_{max}$ are the maximum forward speed and turning
speed of the robot respectively.

\section{The Proposed Method}
\begin{figure}[!htb]
  \centering \includegraphics[width = 0.5\textwidth]{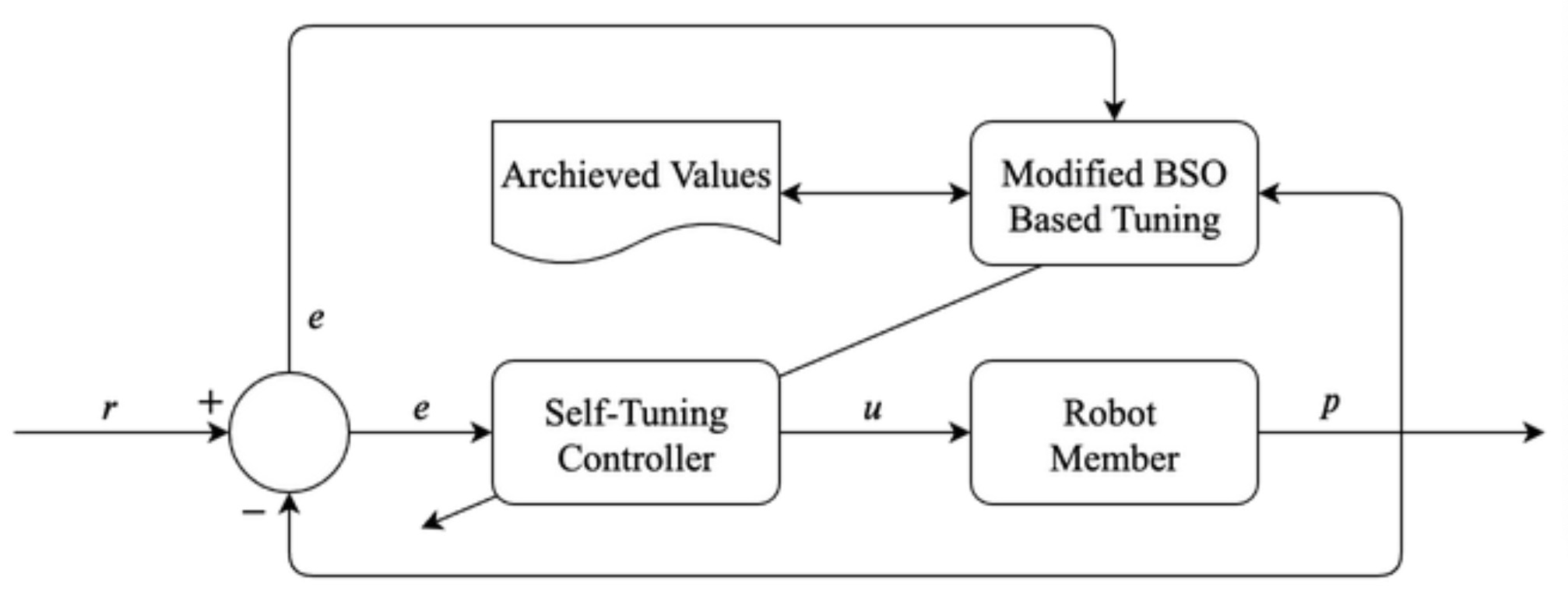}
    \vspace{-2em}
  \caption{Adaptive Control Scheme}
  \label{fig:scheme}
\end{figure}

The proposed control scheme is shown in Figure \ref{fig:scheme}, the
input to the member robot will be calculated by a self-tuning
controller based on the tracking error. Using the objective function
represented in (\ref{eq:4}), we will utilize a modified version of BSO
to optimize the control parameters $\bm{K}_\gamma(k)$ during each step
of movements. Based on the outputs, the current error of each step,
and the archived solutions with corresponding historical errors, the
tuning module will send an optimized set of parameters to the
controller for the next move.
\subsection{Brain Strom Optimization}

The Brain Storm Optimization (BSO) is a relatively new swarm intelligence optimization algorithm, which is inspired by the process of collaborative problem-solving means of human beings, i.e., the brainstorming process. This algorithm can find a relatively optimal solution to a particular problem through convergence and divergence operations within a certain period of time. It generally includes three basic operations, clustering, new solution generation, and selection. The original BSO algorithm uses k-means to cluster in the solution space \cite{shi2011Optimization}. Then it generates new solutions based on the clustering results and selects the better solution in each iteration to retain. Since the k-means clustering is time-consuming, there are many followed efforts to improve the clustering operation. The most representative method is the clustering method in the objective space, namely BSO-OS \cite{shi2015brain}. Each iteration uses every individual's fitness value to divide the entire population into two clusters: elite solutions and normal solutions. Then based on these two clusters, it generates new solutions for the next round of evaluation. Since the fitness value is generally a scalar, this method can significantly increase the calculation speed and can be used in the online PID parameter tuning applications.

The procedure of BSO-OS is shown in Algorithm \ref{alg:bso-os}, in which clustering in the objective space divides the population into two categories: elitists and normals, i.e., the top $perc_e\%$ of the population individuals will be clustered as elitists, and the remaining $(100-perc_e)\%$ will be categorized as normals. Then select one or two individuals to generate a new individual based on the selecting operation. Besides, a disruption operation will be performed on a random individual in the population to increase diversity. This operation is performed by replacing one dimension of a randomly selected individual with a value within a specific range.

\begin{algorithm}[!htb]
  \caption{Procedure of BSO-OS}\label{alg:bso-os}
  \begin{algorithmic}[1]
    \State Population Initialization \While{\sl{not terminated}}
    
    \State Evaluation individuals;

    \State Taking top $perc_e$ percentage as elitists and remaining as
    normals;

    \State Selecting a new individual based on selecting operation

    \State Disrupting a randomly selected individual;
    
    \EndWhile

    \State Updating individuals;
  \end{algorithmic}
\end{algorithm}

Algorithm \ref{alg:select} shows the operation of selecting individuals, where $rand$ is a randomly generated number in $(0,1)$, $p_e$ is the probability to use elitists not normals to generate a new individual, $p_{one}$ is the probability to generate a new individual based on one selected individual rather than two selected individuals.

\begin{algorithm}[!htb]
  \caption{Procedure of Selecting Individuals}\label{alg:select}
  \begin{algorithmic}[1]

    \If{$rand< p_{e}$} \Comment{generate a new individual based on
      elitists}

    \If {$rand < p_{one}$}

    \State Generate a new individual based on one randomly selected
    elitist;

    \Else

    \State Generate a new individual based on two randomly selected
    elitists; \EndIf

    \Else \Comment{generate a new individual based on normals}

    \If {$rand < p_{one}$}
    
    \State Generate a new individual based on one randomly selected
    normal;

    \Else

    \State Generate a new individual based on two randomly selected
    normals; \EndIf \EndIf
  \end{algorithmic}
\end{algorithm}

A new solution is realized by adding the a Gaussian random number to the selected solution, according to Eq.(\ref{eq:genNew}).
\begin{equation}
  \label{eq:genNew}
  x_{new}^i(t+1) = x_{selected}^i+random(t)\xi(t)
\end{equation}
where $x_{new}^i$ is the $i$th dimension of the newly created
individual, and $x_{selected}^i$ is the corresponding dimension of the
selected individual based on Algorithm \ref{alg:select}, which can be
determined by Eq.(\ref{eq:select}), and $random()$ is a random value
that obeys the Gaussian distribution.
\begin{equation}
  \label{eq:select}
  \begin{aligned}
    &x_{selected}^i(t) = \\
    &\begin{cases}
      x_{old1}^i(t), \qquad rand < p_e \\
      rand(t) x_{old1}^i(t) + (1-rand(t))x_{old2}^i(t),
      \text{Others}
    \end{cases}
  \end{aligned}
\end{equation}
where $x_{old1}$ and $x_{old2}$ represent two individuals selected from the current population. $rand()$ returns a random value in the range of $(0,1)$, and the $\xi(t)$ is the step size, which relevant to the predefined total iteration numbers, can be determined by (\ref{eq:xi}).
\begin{equation}
  \label{eq:xi}
  \xi(t) = logsig(\frac{\frac{T}{2}-t}{k})r(t)
\end{equation}
where $logsig()$ is a logarithmic sigmoid transfer function, $T$ is the predefined maximum number of iterations, $t$ is the current iteration number, $k$ is for changing $logsig()$ function's slope, and $r(t)$ is a random value within $(0,1)$.

\subsection{Modified BSO Based Tracking Control}
The proposed BSO based online tuning PID control for each member is
shown in Algorithm \ref{alg:motionControl}. The inputs are sensor data
from the robot detector, which can indicate the range and bearing of
other robots in its field of view. The predefined parameters include
safe distance for anti-collision, and parameters for BSO, such as the
elite percentage, and probability values defined in the algorithm. By
keeping an elite and normal list, the robot controller updates the PID
parameters in every step and gets the relatively optimal value for the
next move. It needs to be mentioned that the target pose may change
during the movements. The present method is able to track changing
targets with relatively optimal control parameters.

\begin{algorithm}[!htb]
  \caption{Adaptive Coordinated Motion Control}\label{alg:motionControl}
  \begin{algorithmic}[1]
    \State {\bf Input:} Robot Detections, Obstacle Detections

    \State Initialize arguments: Safe Distance ($d_s$), Population
    Size ($N$), $perc_e$, $p_e$, and $p_{one}$;

    \State Randomly Generate $N$ Solutions;

    \State Evaluate the Generated $N$ Solutions;

    \State Initialize Elite List and Normal List.

    \While {True}

    \If{$rand< p_{e}$} \Comment{new solution from based on
      elitists}

    \If {$rand < p_{one}$}

    \State Generate a new individual based on one randomly selected
    solution from elite list;

    \Else

    \State Generate a new individual based on two randomly selected
    solutions from elite list; \EndIf

    \Else \Comment{generate a new individual based on normals}

    \If {$rand < p_{one}$}
    
    \State Generate a new individual based on one randomly selected
    solution from normal list;

    \Else

    \State Generate a new individual based on two randomly selected
    solutions from normal list;

    \EndIf
    \EndIf

    \State Evaluate the Generated New Solution;

    \State Update the Elite and Normal Lists;
    
    \State $\bm{K}_{pid}^*=$ The best solution in Elist;
    
    \State Follow the reference using (9);

    \EndWhile
  \end{algorithmic}
\end{algorithm}

\section{Results}
\subsection{Configuration}
\begin{figure}[!htb]
  \centering \includegraphics[trim = 100 200 100 200, clip, width =
  0.4\textwidth]{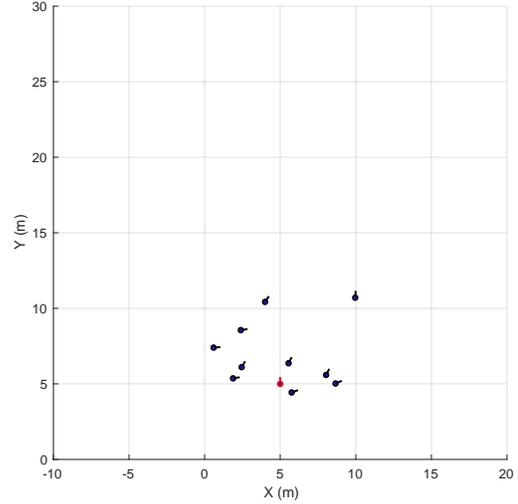} 
  \caption{Initial Configuration}
  \label{fig:init}
\end{figure}

The sensor range of a single robot is set to 10m, and the safe
distance $d_s$ is set to 0.5m. The desired distance and bearing angle
to the target robot is 1m and $\pm \pi/4$, respectively. For the left
part of a v-shaped formation, this angle is $\pi/4$, while the right
part is $\pi/4$ correspondingly. Initially, the swarm members are
randomly distributed in an $N \times N$m area, and a swarm leader is
assigned to guide the whole swarm to follow a specific trajectory. The
position of the swarm leader is located at $((N-1)/2,(N-1)/2)$, where
$N$ is the total number of the swarm members. We set $N$ to be odd for
convenience. For example, as shown in Figure \ref{fig:init}, a swarm
with 11 members is randomly distributed in an area of $30 \times 30$,
the leader for the whole swarm is located at $(5, 5)$. The Mobile
Robot Simulation Toolbox for Matlab is used for the verification of
the proposed method. All the tests were implemented on an iMac with
3.6 GHz Intel Core i9, 8GB DDR4 memory with Matlab 2019b.

\subsection{Formation Control}
\begin{figure}[!htb]
  \centering \includegraphics[trim = 100 200 100 200, clip, width =
  0.4\textwidth]{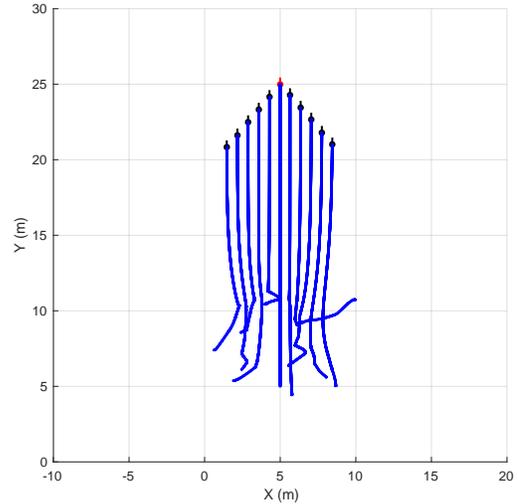}
  \caption{Formation trajectories of with followers}
  \label{fig:trace}
\end{figure}

We first tested the proposed formation control method under the above
initial distribution. The leader is configured to move from $(5,5)$ to
$(5,25)$, the trajectories, and the final formation of the swarm are
shown in Figure \ref{fig:trace}.

\begin{figure}[!htb]
  \centering \includegraphics[trim=50 200 50 200, clip, width =
  0.22\textwidth]{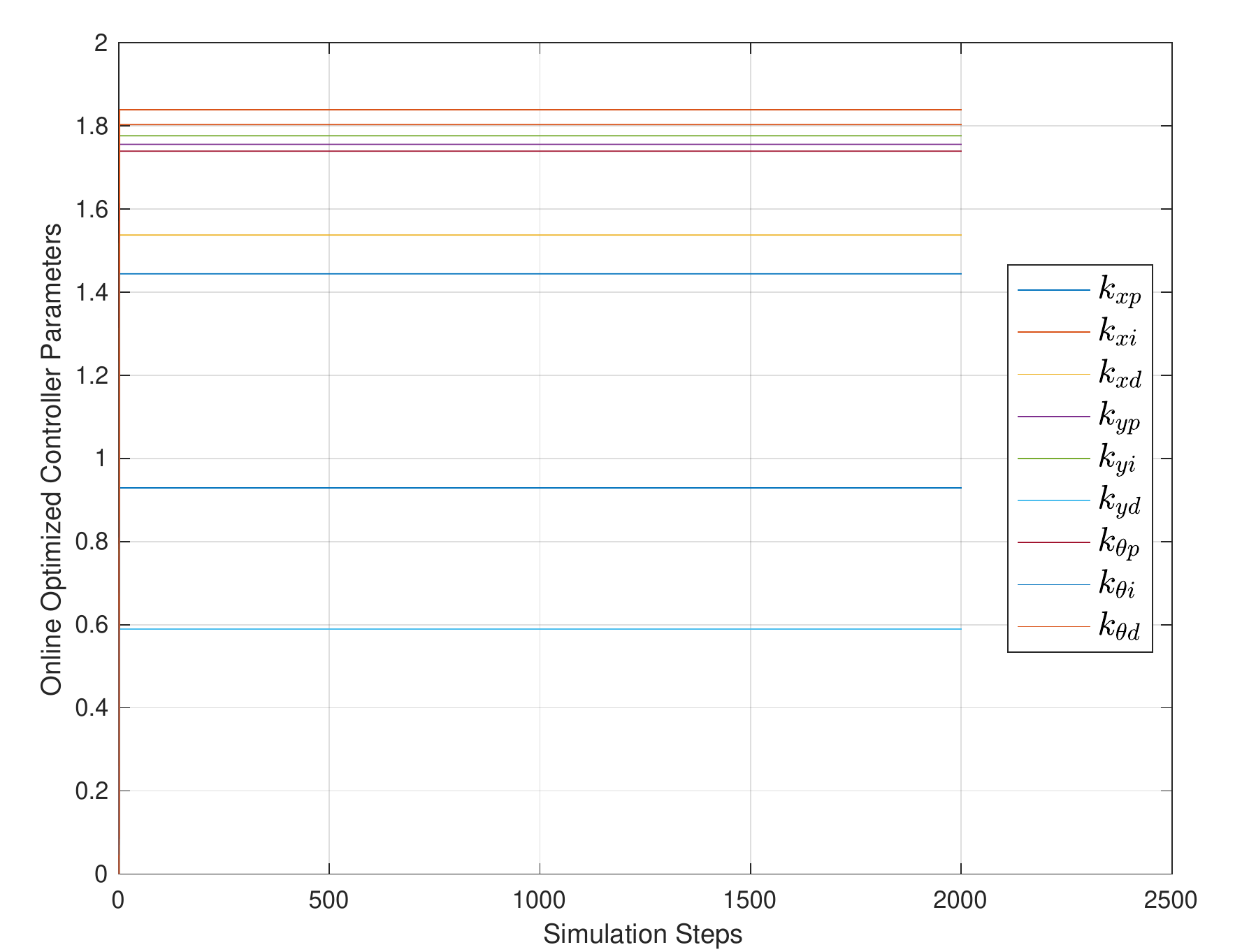} \includegraphics[trim=50
  200 50 200, clip,width = 0.22\textwidth]{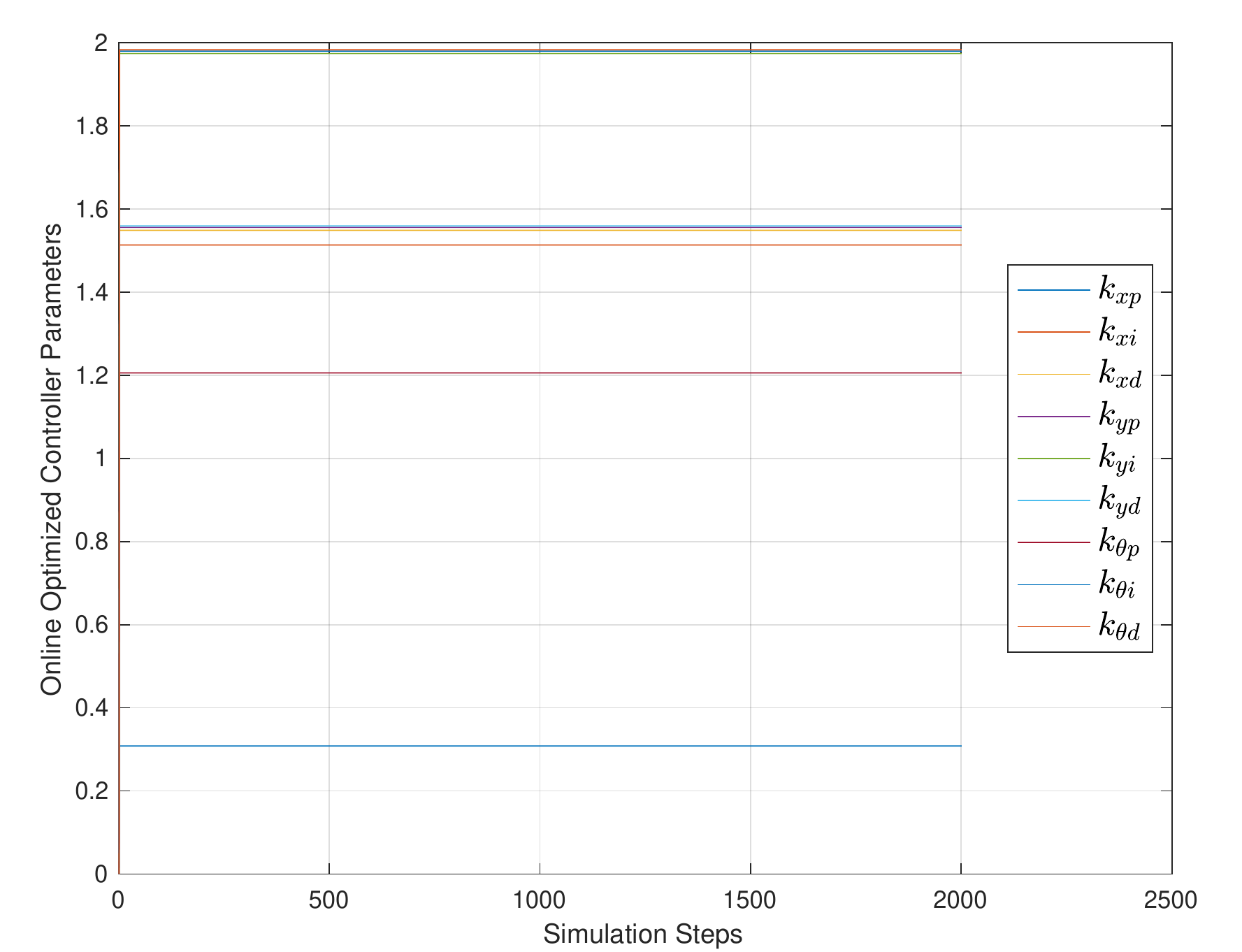}
  \includegraphics[trim=50 200 50 200, clip, width =
  0.22\textwidth]{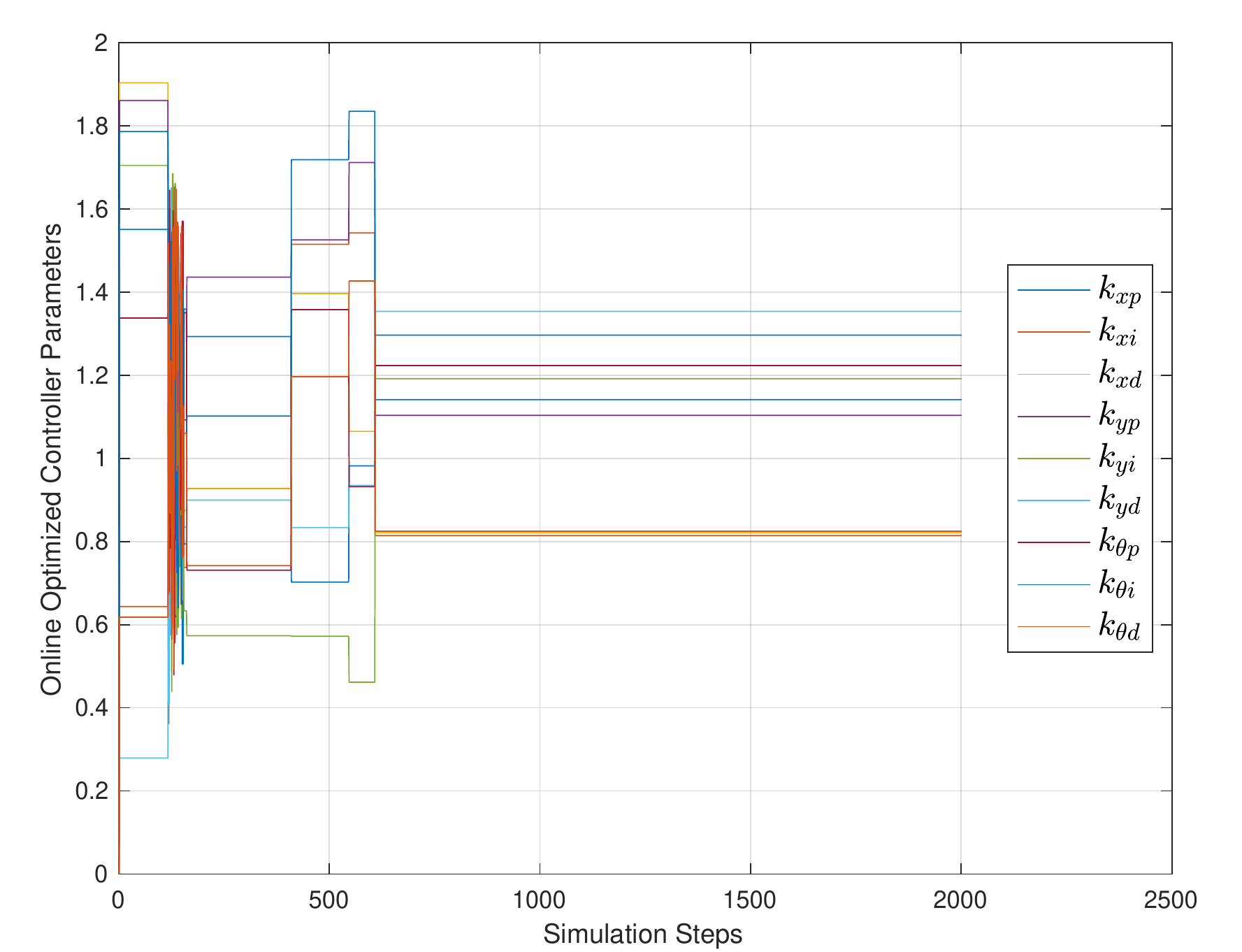} \includegraphics[trim=50
  200 50 200, clip, width = 0.22\textwidth]{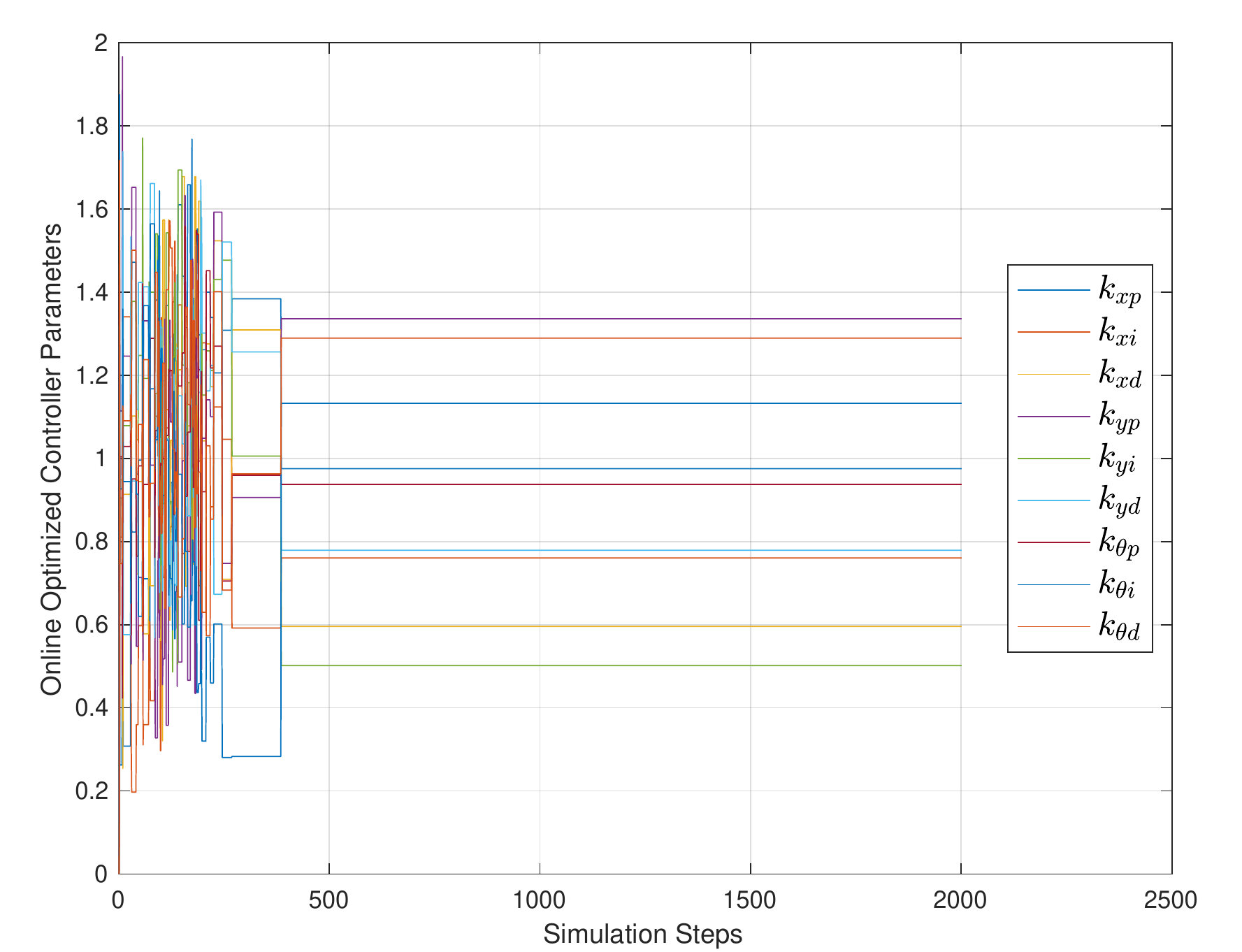}
  \includegraphics[trim=50 200 50 200, clip, width =
  0.22\textwidth]{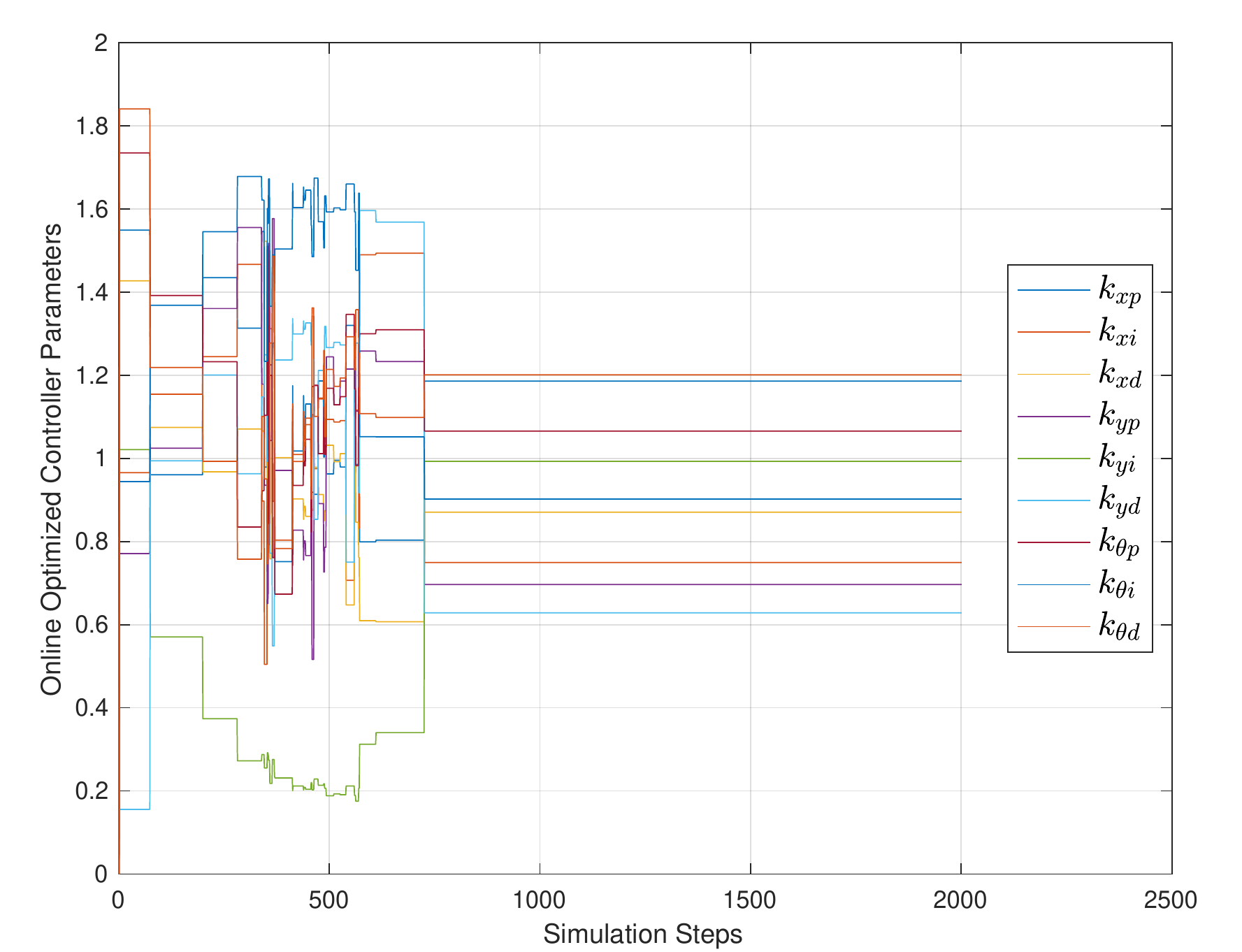} \includegraphics[trim=50
  200 50 200, clip, width = 0.22\textwidth]{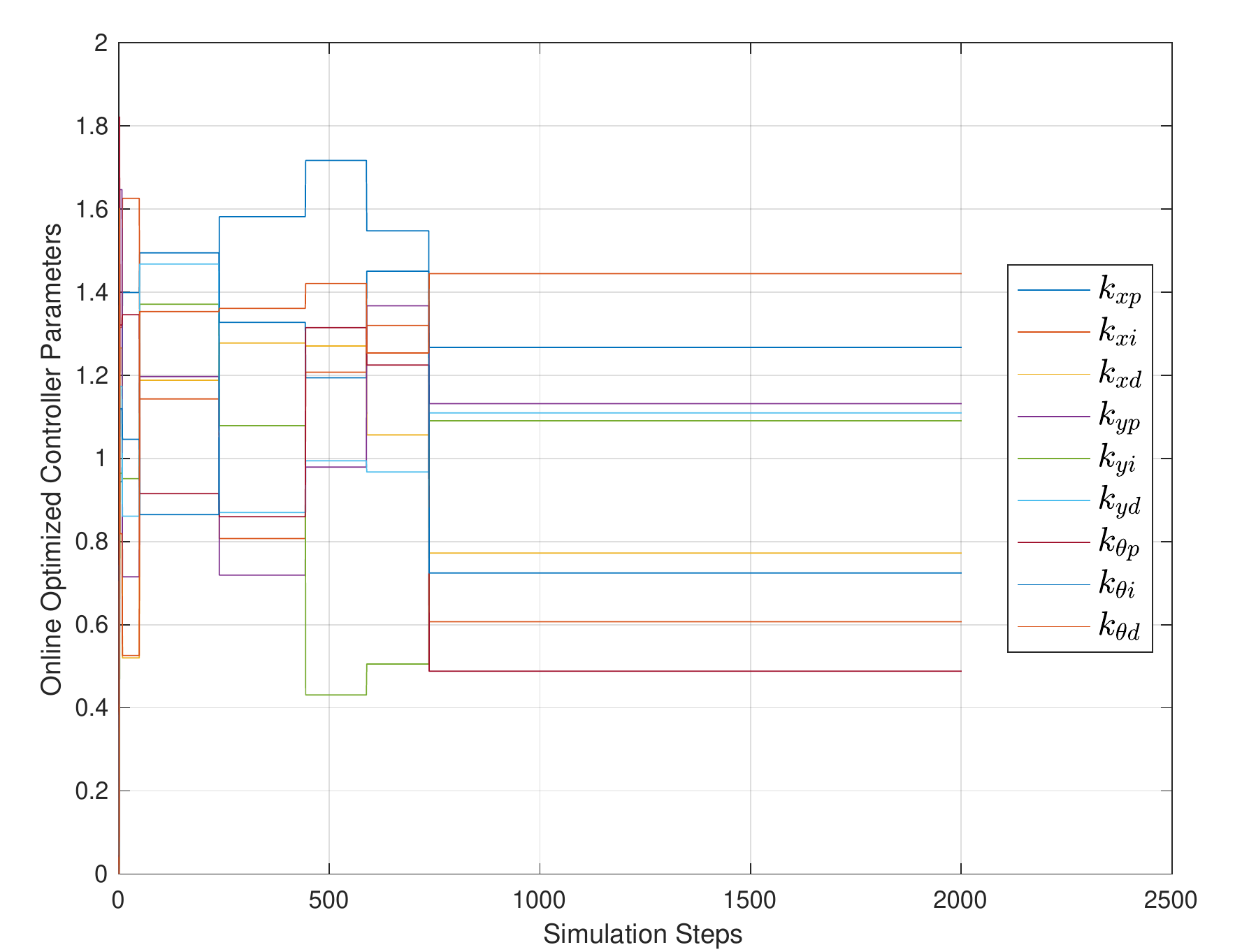}
  \includegraphics[trim=50 200 50 200, clip,width =
  0.22\textwidth]{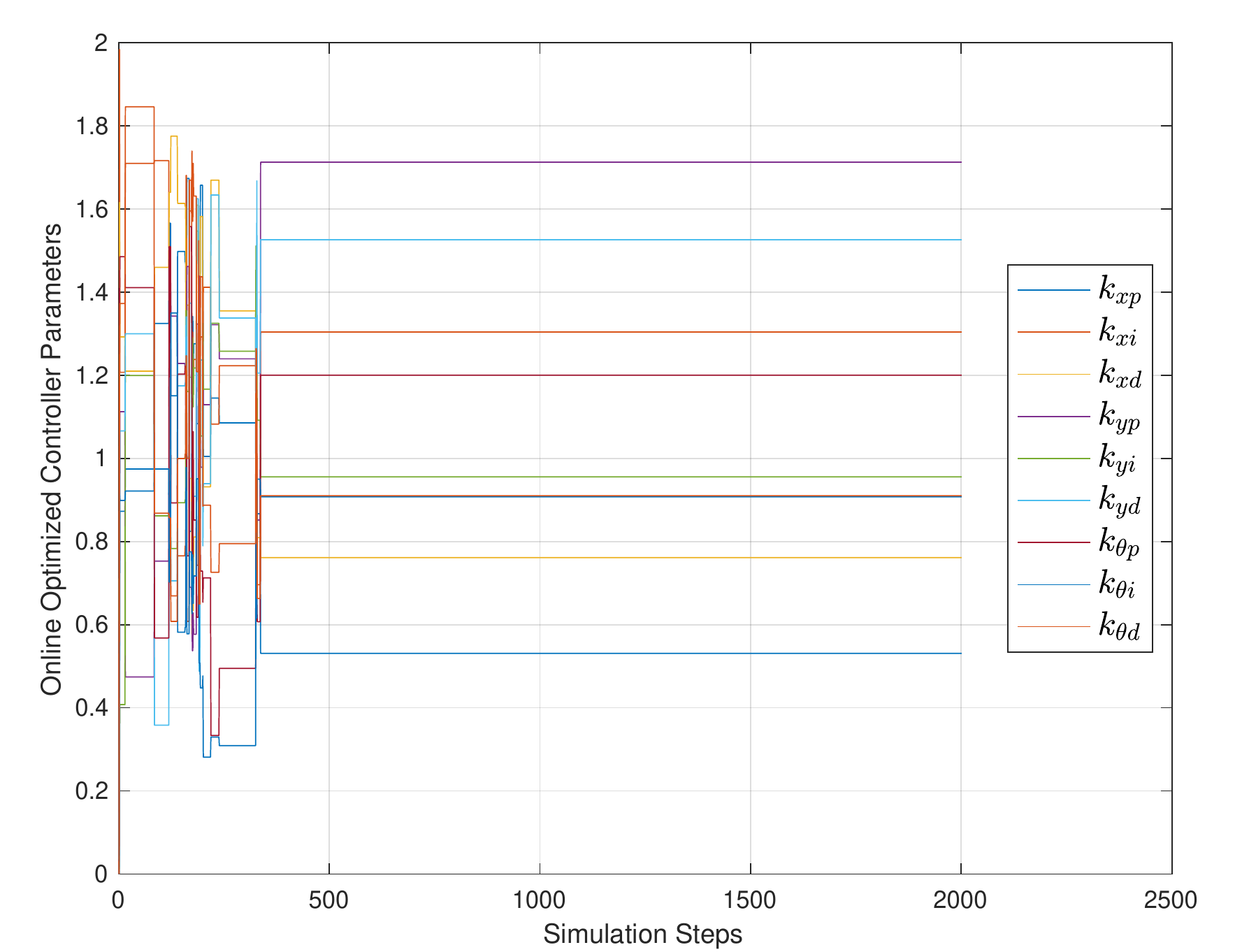} \includegraphics[trim=50
  200 50 200, clip, width = 0.22\textwidth]{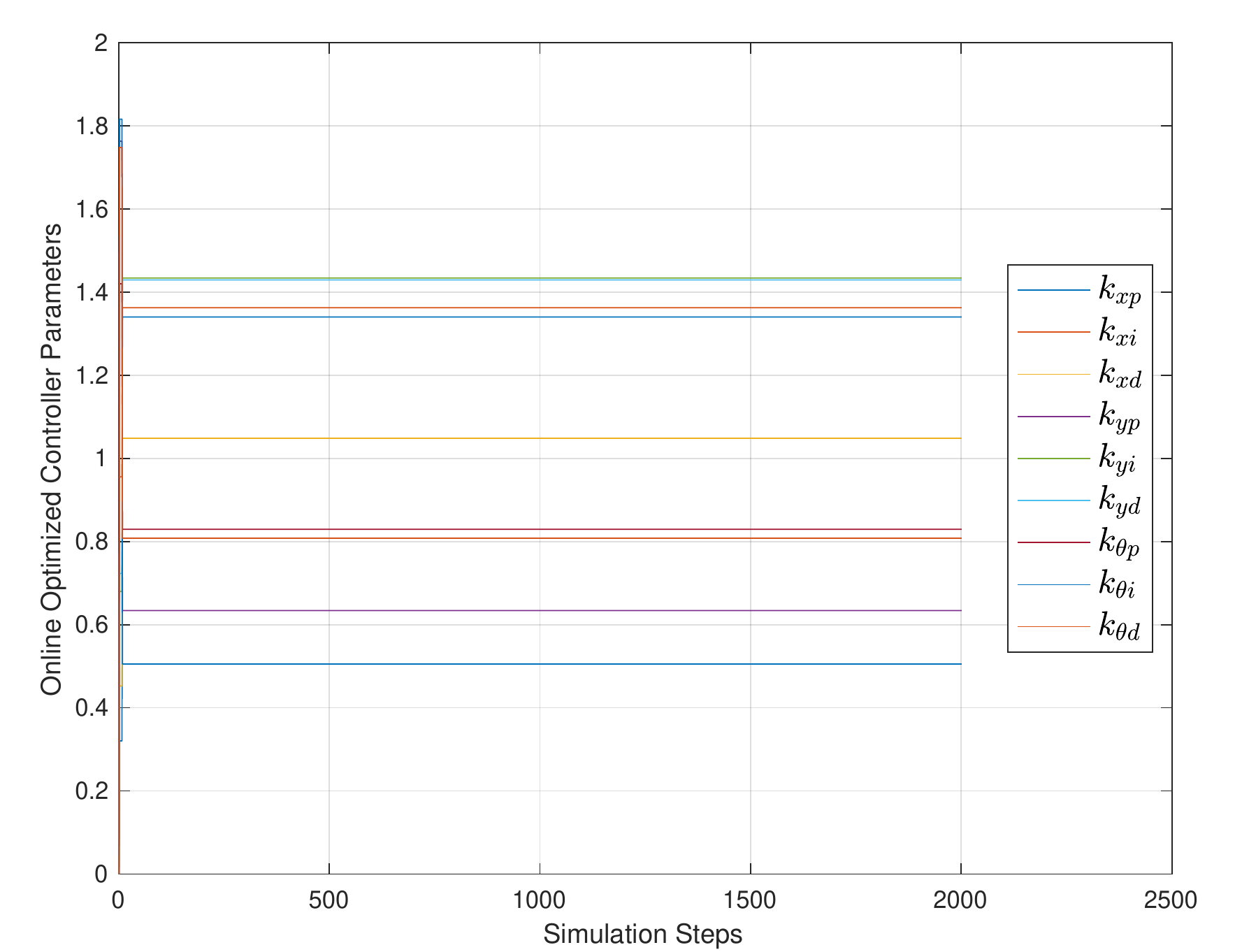}
  \includegraphics[trim=50 200 50 200, clip, width =
  0.22\textwidth]{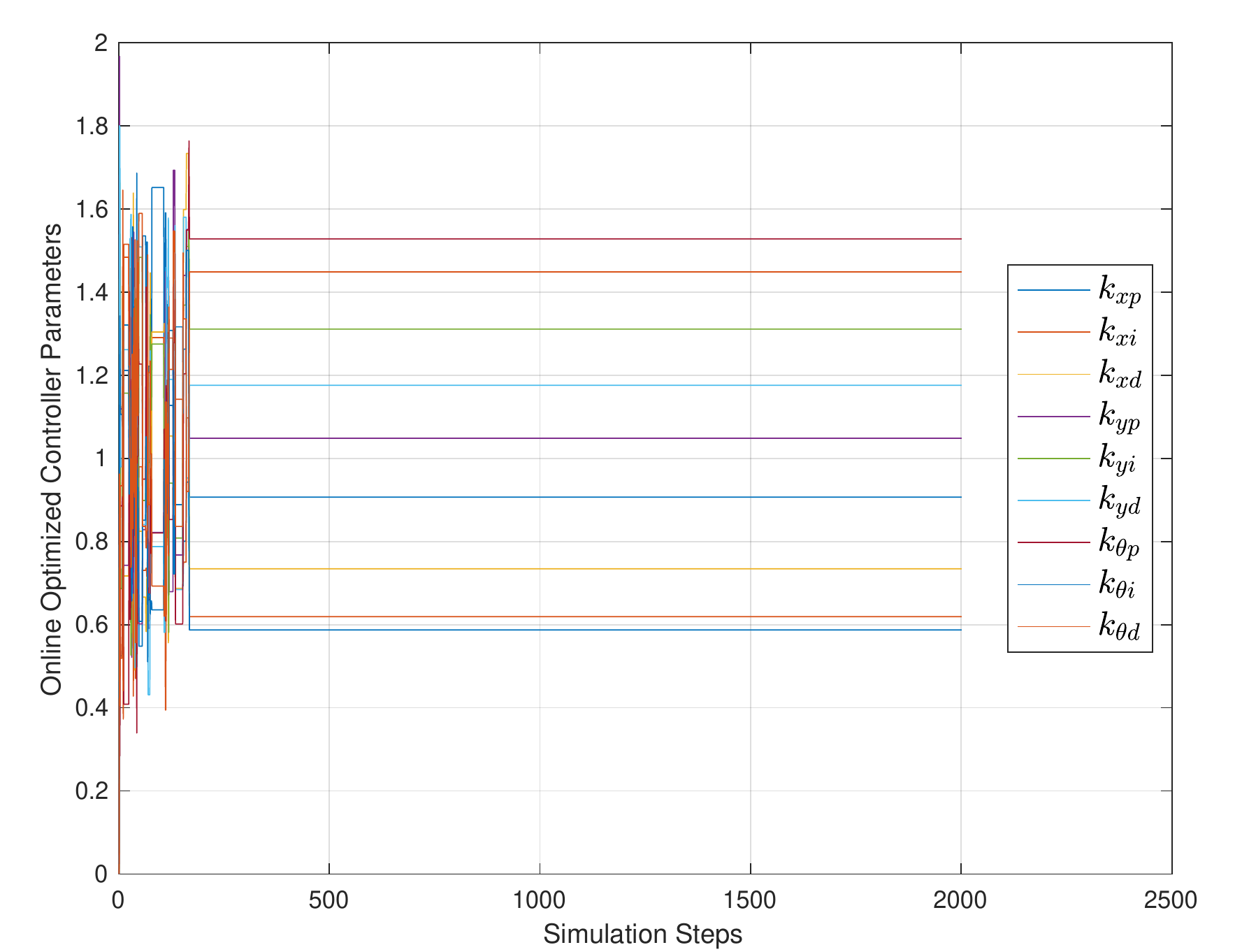} \includegraphics[trim=50
  200 50 200, clip, width = 0.22\textwidth]{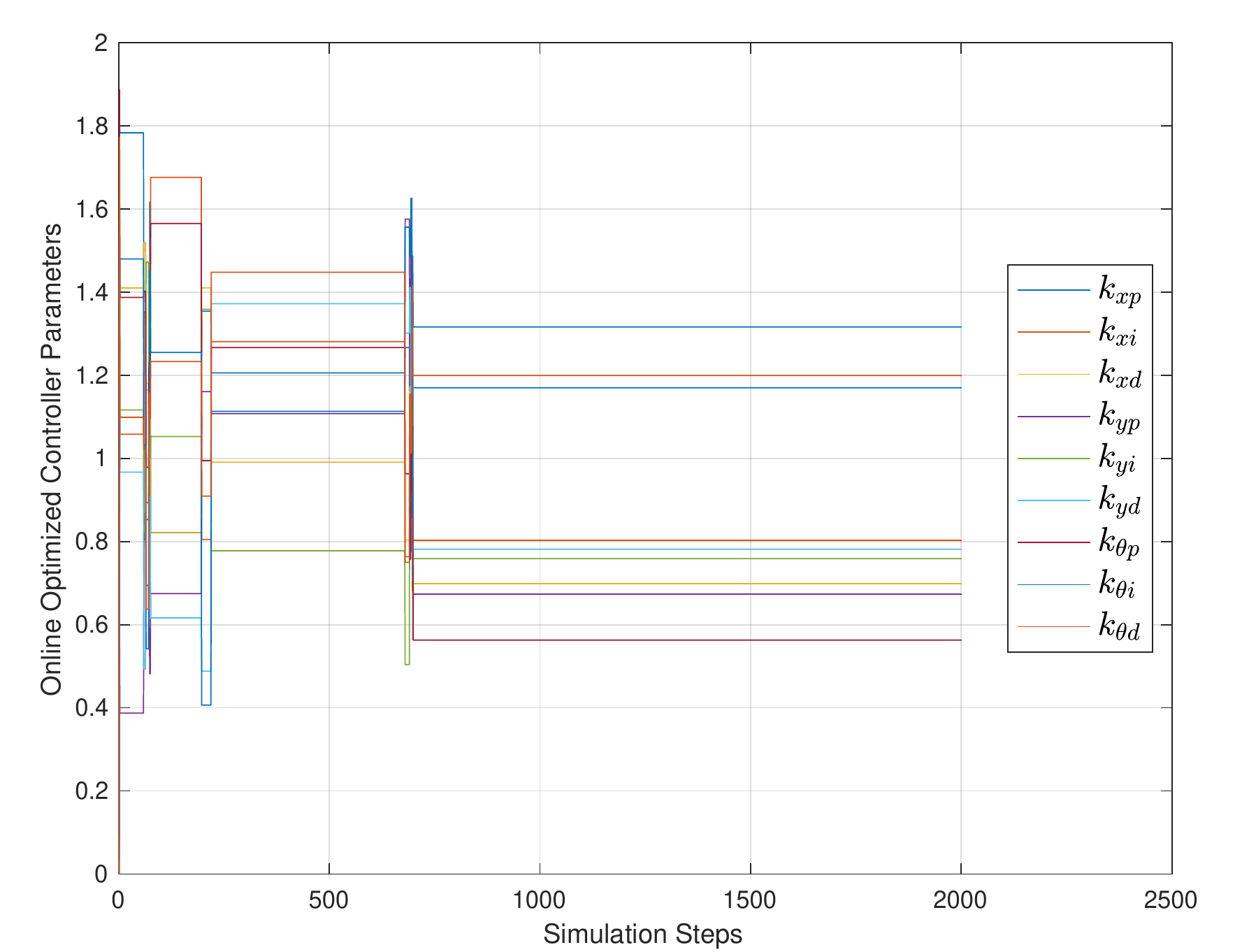}
  \caption{The control parameters convergence process of each member}\label{fig:optK}
\end{figure}

It should be noted that the swarm leader here is adopted only to guide
the whole swarm, but not every member will follow it. Each follower
will track the nearest neighbor in a particular direction. The
parameters' changes for each follower are shown in Figure
\ref{fig:optK}. The control parameters for the PID tracking controller
for each member is optimized during iterations. From those curves, we
know that some of the controller parameters converged after a few
iterations, such as for robots 1, 2, and 8, while the controller
parameters of other robots converged after certain steps of guided
searching operations. The simulation indicated that the proposed
method can determine the controller parameters with optimized values.

Furthermore, the tracking errors of each robot in the swarm are also
converged to satisfactory values, as shown in Figure 6. Due to the
followed target may change during the formation procedure, or because
of the anti-collision operations, the errors curve may have some
vibrations, but after the formation is formed and keep stable, the
errors will decrease and converge to minimal values.
\begin{figure}[!htb]
  \centering \includegraphics[trim = 20 200 20 200, clip, width =
  0.4\textwidth]{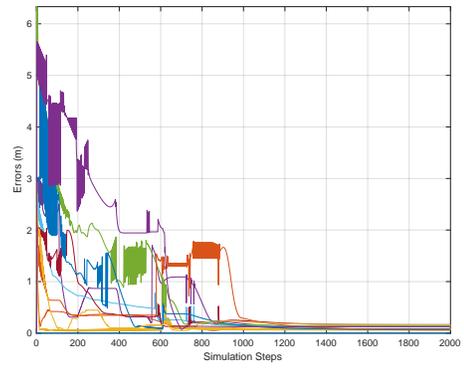}
  \caption{Tracking Errors of each robot}
  \label{fig:Error}
\end{figure}

\subsection{Flexibility}
\begin{figure}[!htb]
  \centering \subfloat[Initial Configuration\label{fig:init2}]{%
    \includegraphics[trim = 100 200 100 200, clip, width =
    0.4\textwidth]{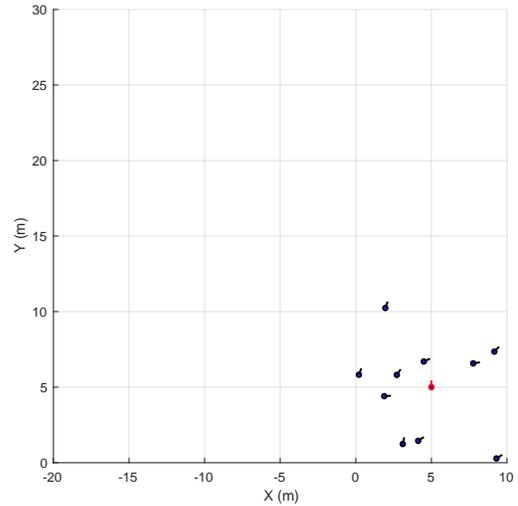}}\\ \vspace{-1em}
  \subfloat[Final formation and traces \label{fig:trace2}]{%
    \includegraphics[trim = 100 200 100 200, clip, width =
    0.4\textwidth]{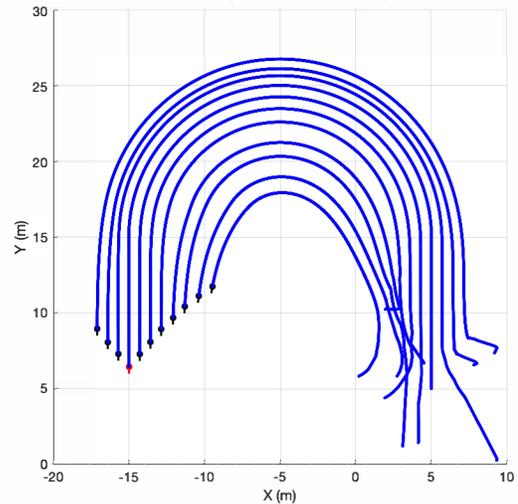}}
  \caption{Results of Flexibility Simulation}
  \label{fig:flexibility}
\end{figure}

We also tested the flexibility of the proposed method by letting the
swarm leader perform a U-turn. The initial distribution and
trajectories of the swarm are shown in \ref{fig:init2} and Figure
\ref{fig:trace2}, respectively.

\begin{figure}[!htb]
  \centering \includegraphics[trim=50 200 50 200, clip, width =
  0.22\textwidth]{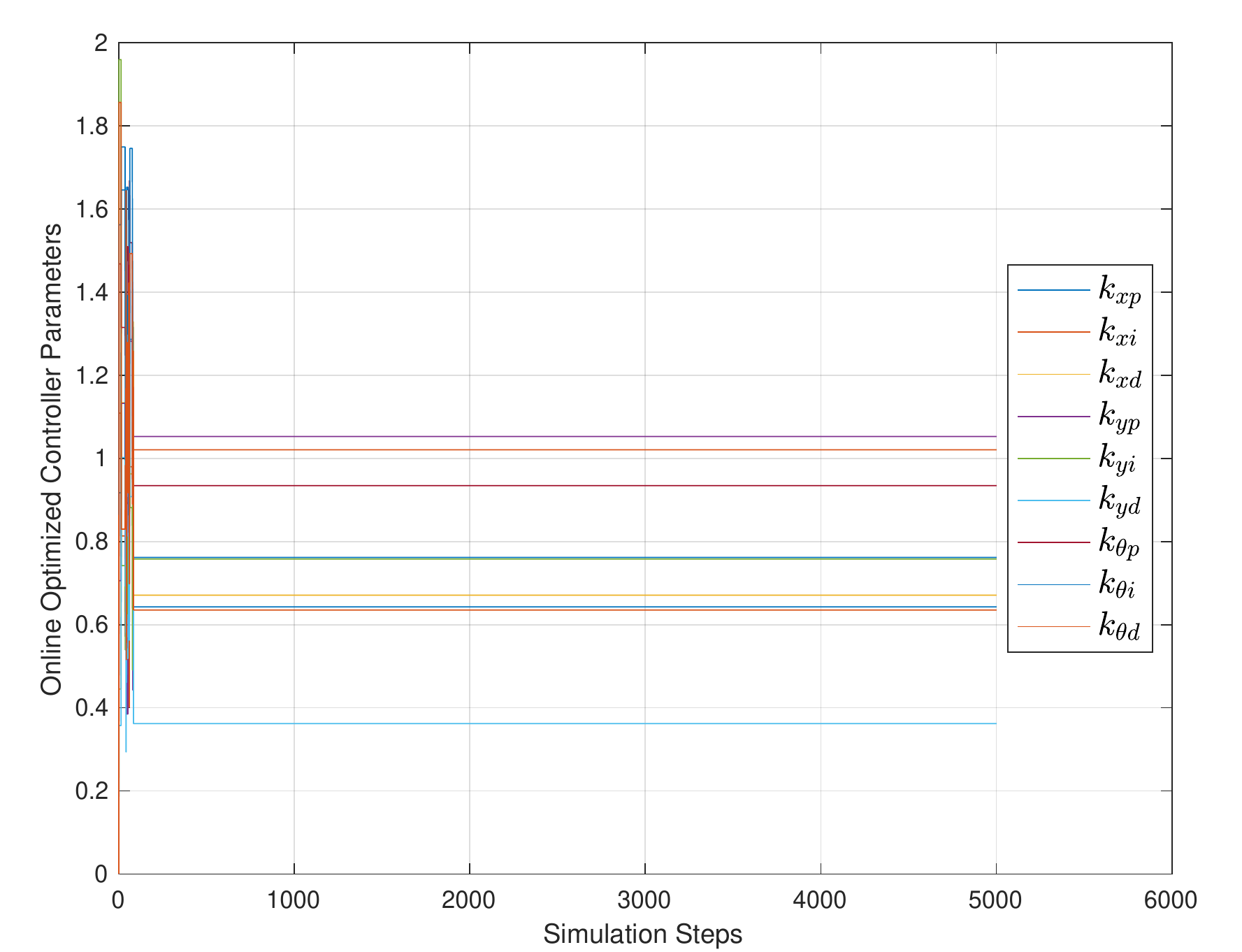} \includegraphics[trim=50 200
  50 200, clip,width = 0.22\textwidth]{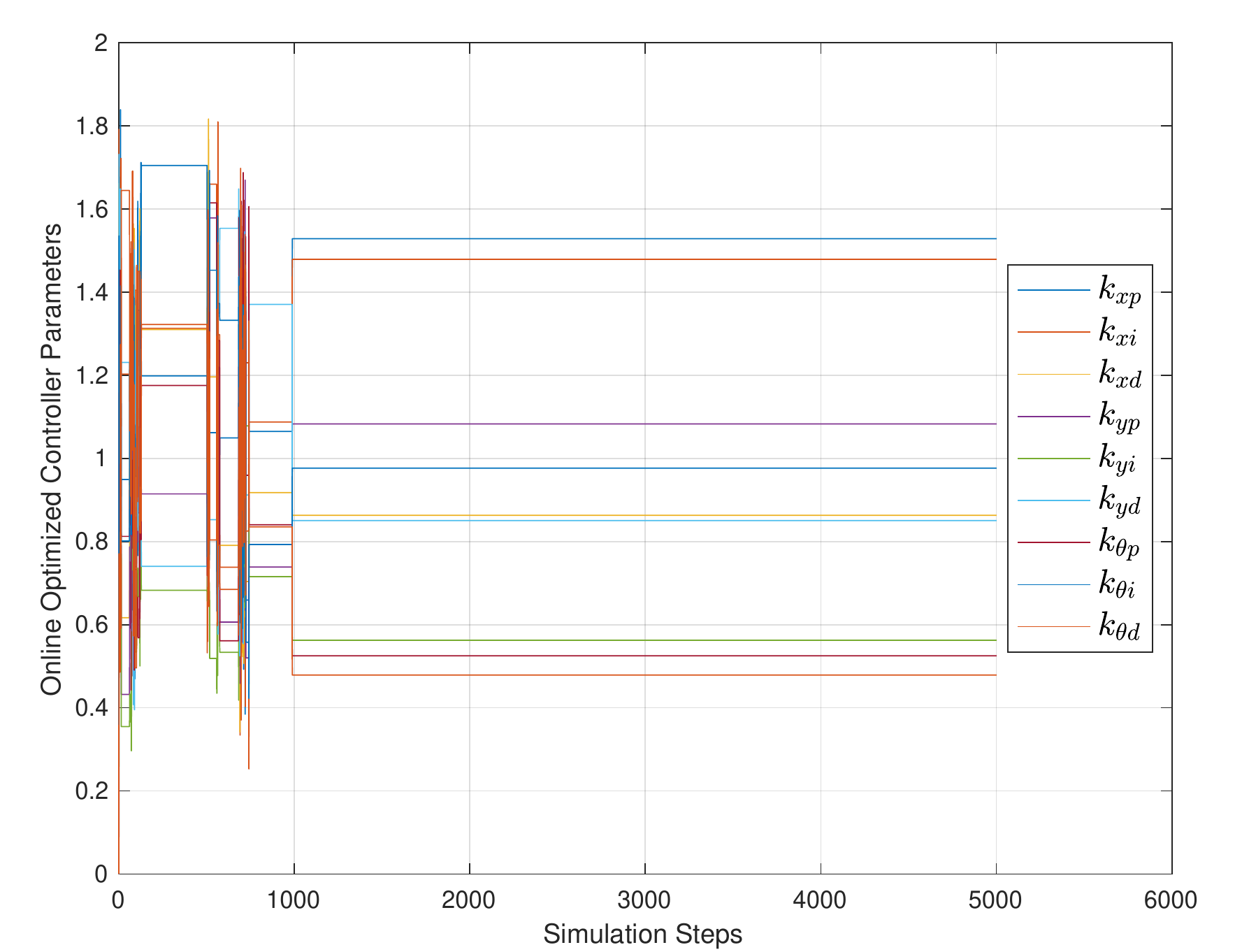}
  \includegraphics[trim=50 200 50 200, clip, width =
  0.22\textwidth]{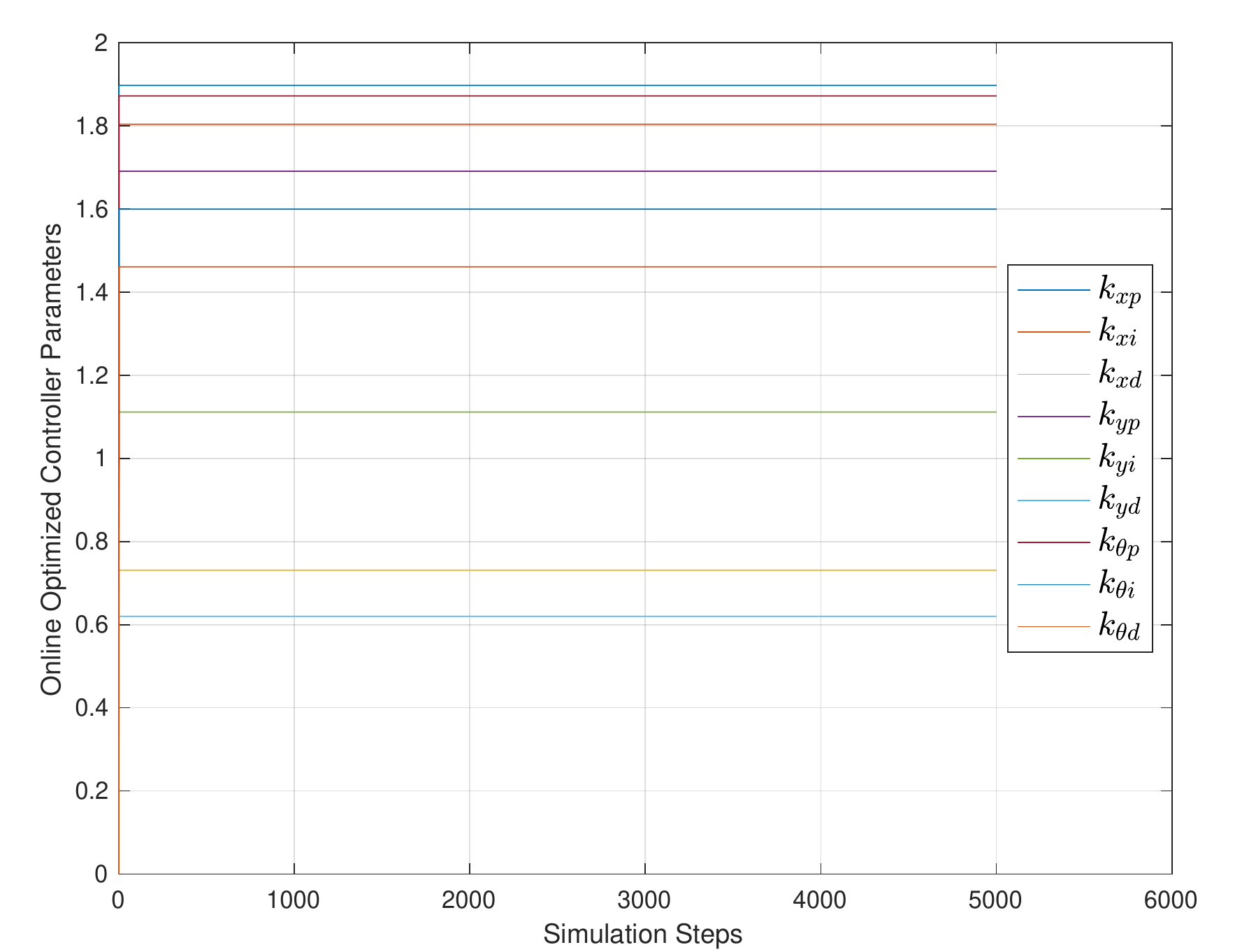} \includegraphics[trim=50 200
  50 200, clip, width = 0.22\textwidth]{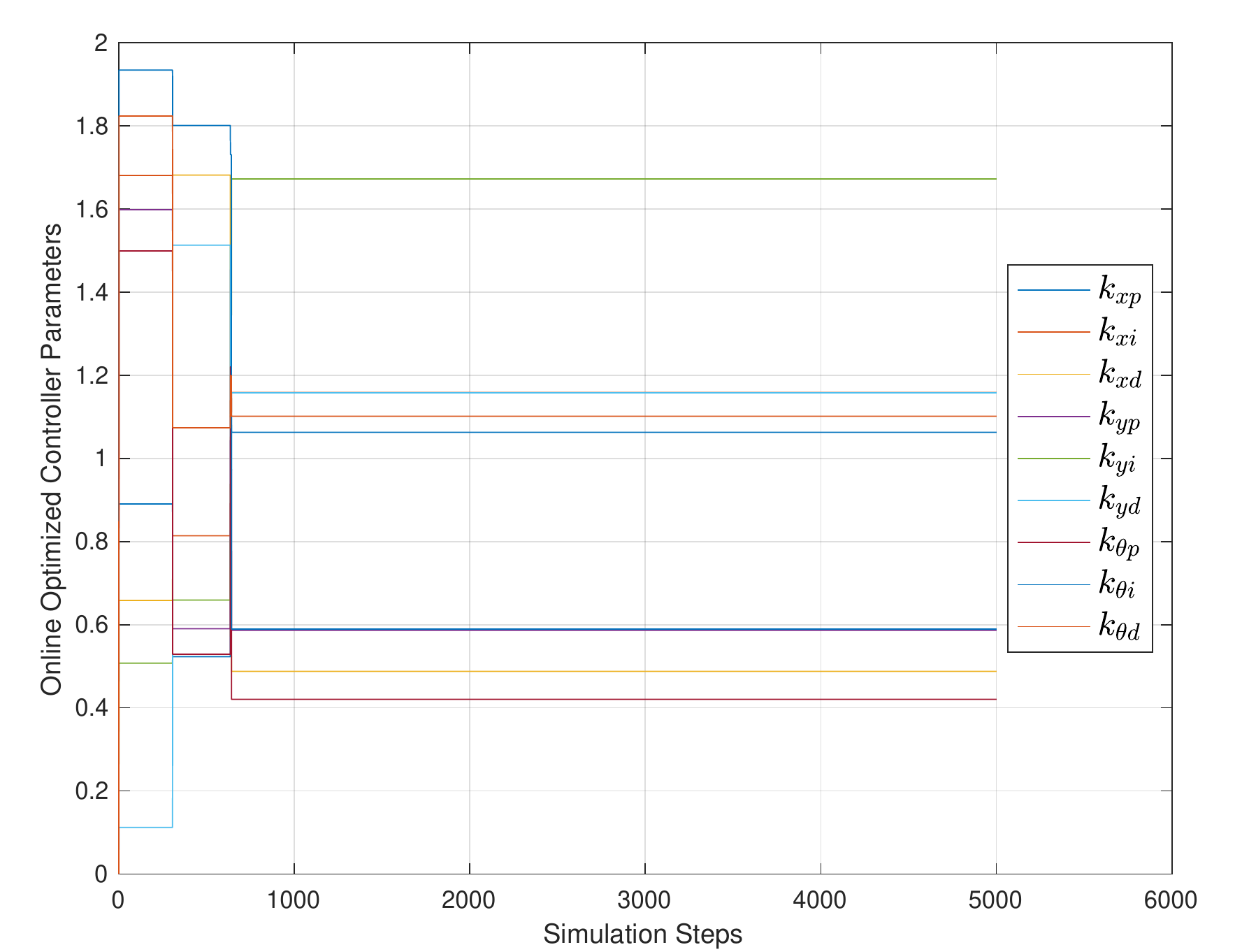}
  \includegraphics[trim=50 200 50 200, clip, width =
  0.22\textwidth]{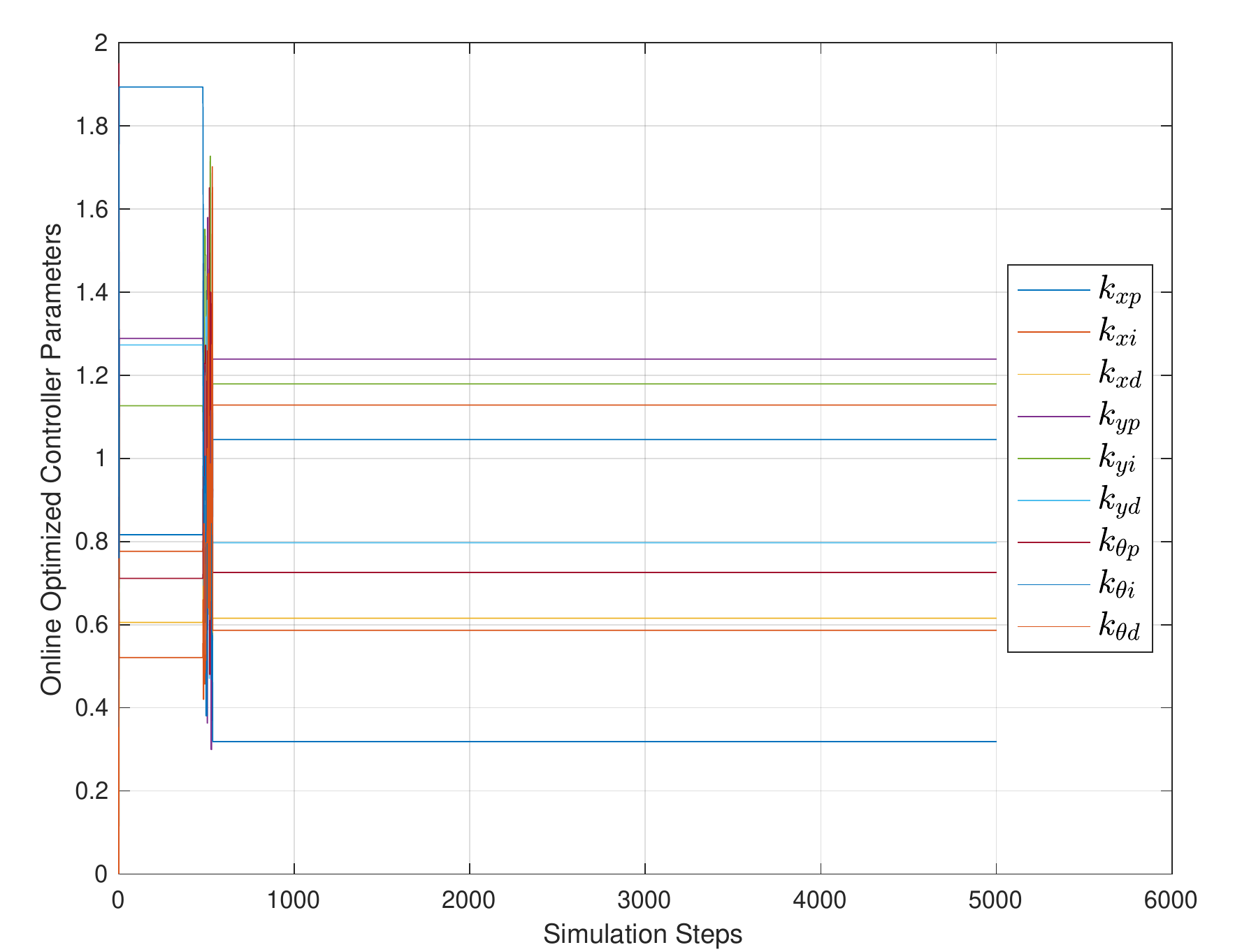} \includegraphics[trim=50 200
  50 200, clip, width = 0.22\textwidth]{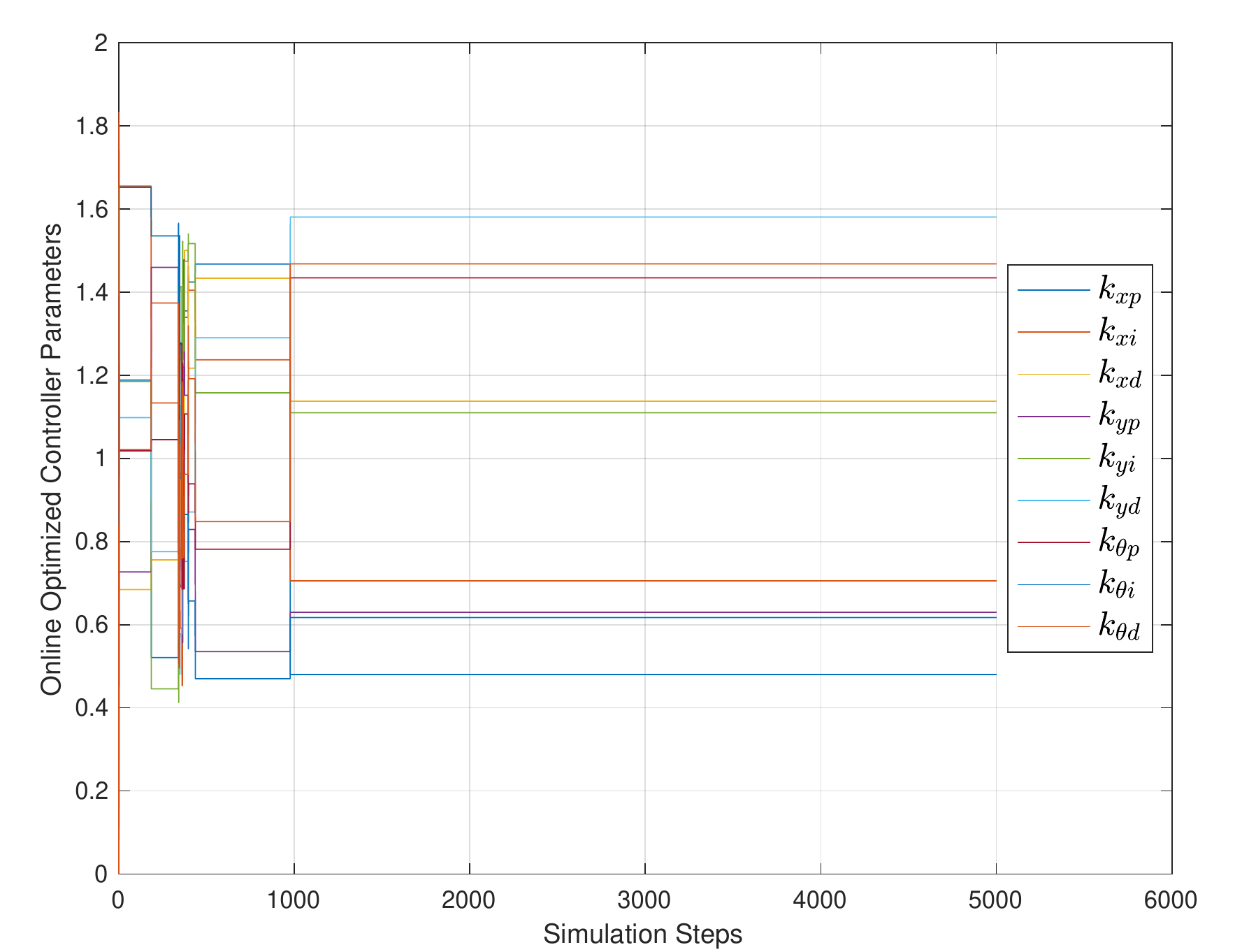}
  \includegraphics[trim=50 200 50 200, clip,width =
  0.22\textwidth]{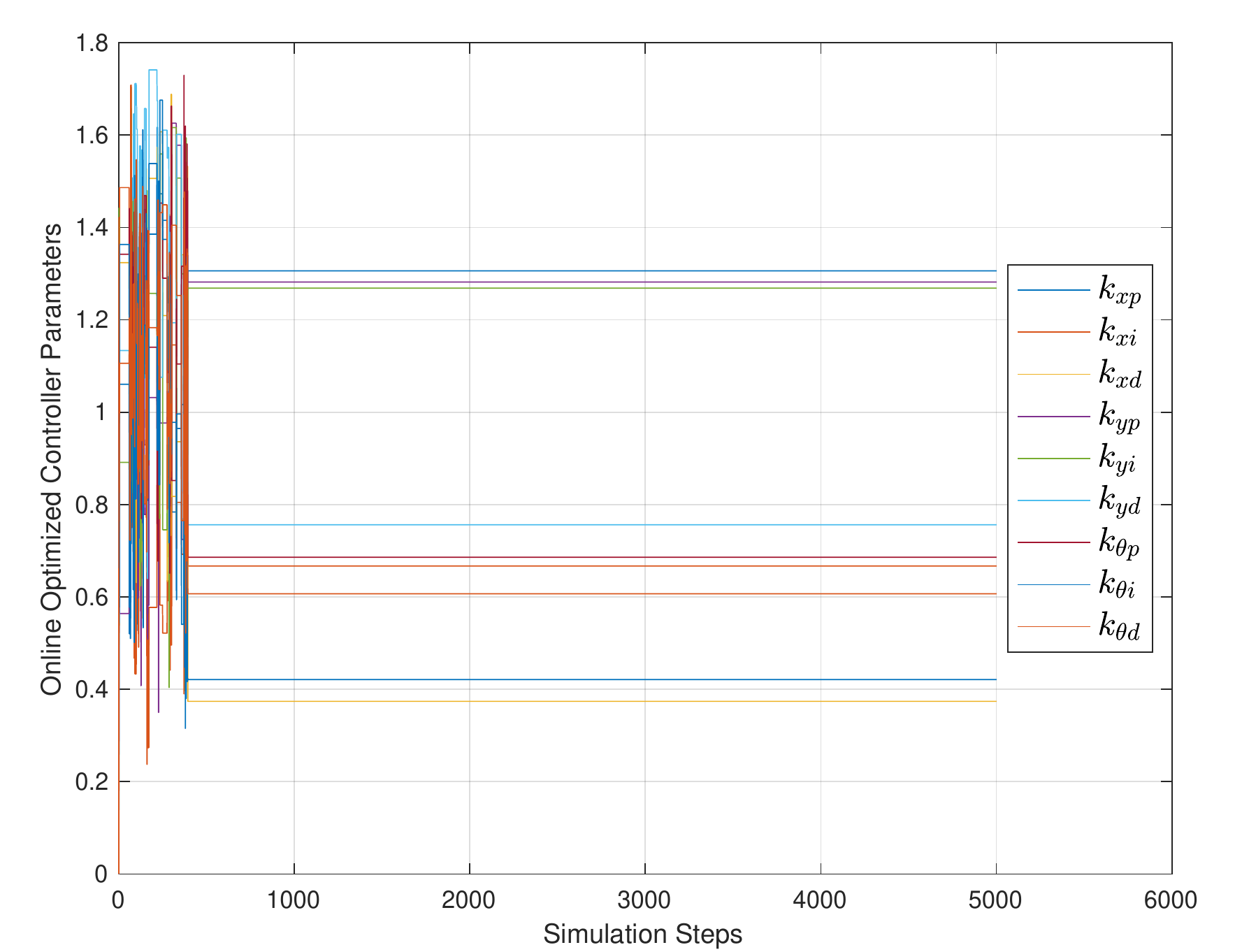} \includegraphics[trim=50 200
  50 200, clip, width = 0.22\textwidth]{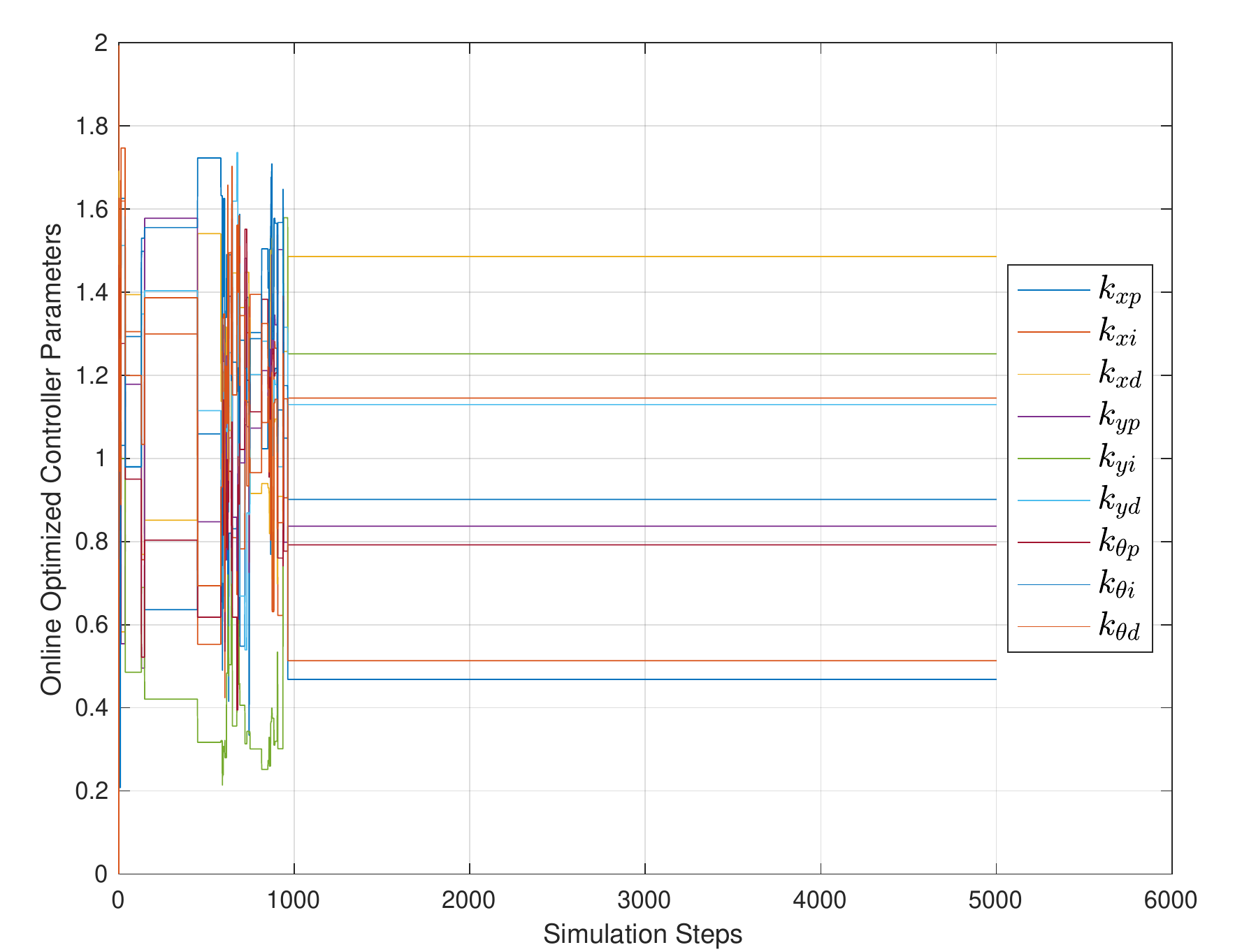}
  \includegraphics[trim=50 200 50 200, clip, width =
  0.22\textwidth]{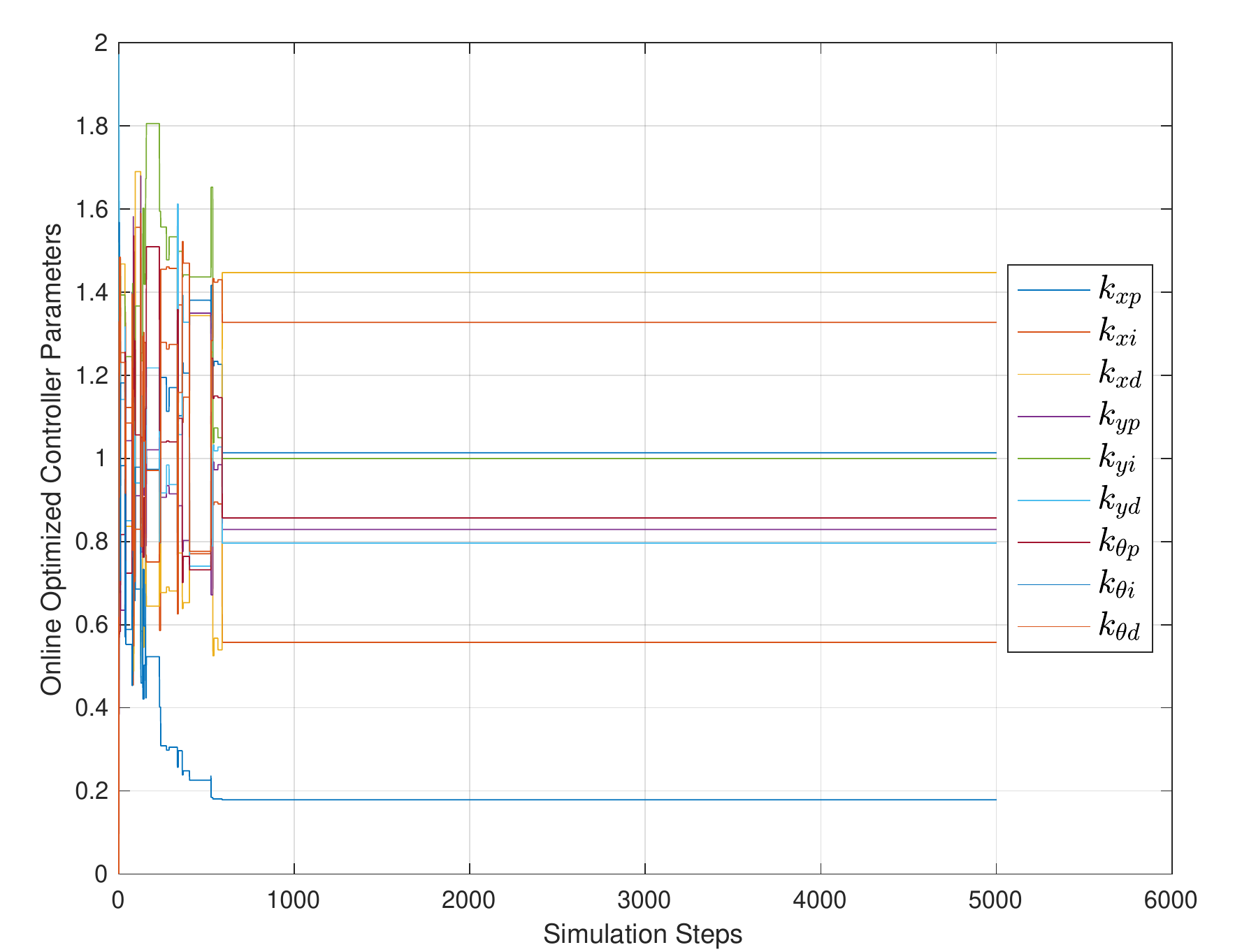} \includegraphics[trim=50 200
  50 200, clip, width = 0.22\textwidth]{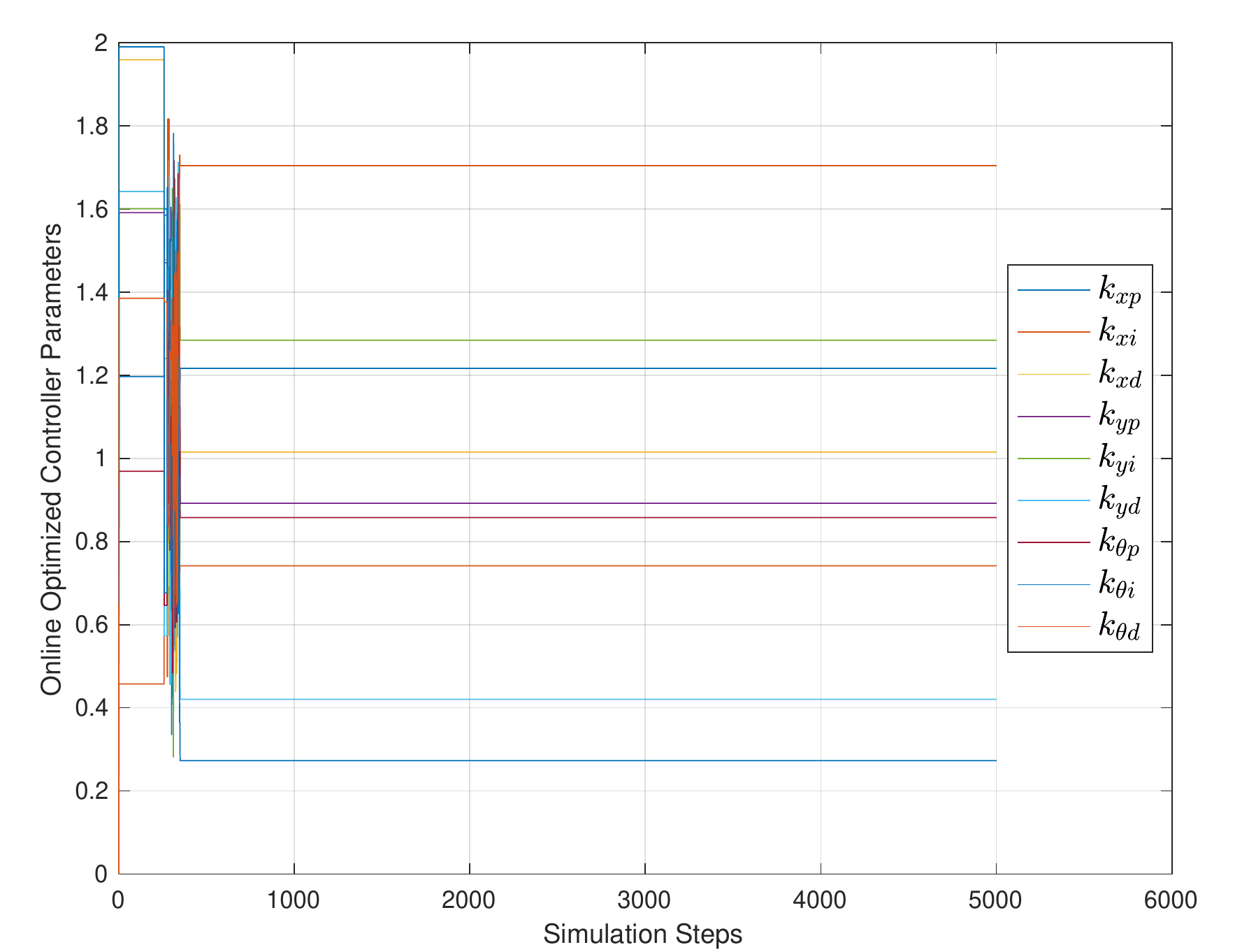}
  \caption{The control parameters convergence process of each member}\label{fig:optK2}
\end{figure}

\begin{figure}[!h]
  \centering \includegraphics[trim = 20 200 20 200, clip, width =
  0.4\textwidth]{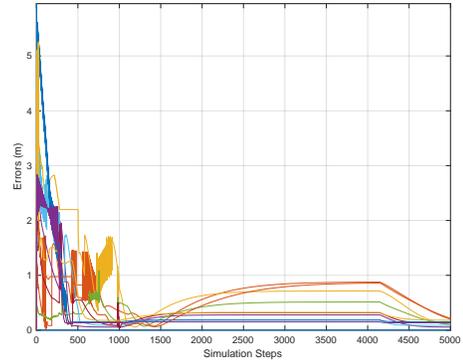}
  \caption{Errors changing during simulation}
  \label{fig:Error2}
\end{figure}

We can see from the figures, although the leader did not broadcast its
velocity to others, the proposed method is able to follow the target
member to form a predefined formation, with the online tuning optimal
controller. In this simulation, the parameters reach the convergent
values after the maximum of 1000 simulation steps, which are shown in
Figure \ref{fig:optK2}. The corresponding errors' changes are given in
Figure \ref{fig:Error2}. Since we constrained the maximum forward and
turning speed of each robot, the errors were increasing slightly when
the whole swarm was making a U-turn.

\subsection{Scalability}
\begin{figure}[!htb]
  \centering \includegraphics[trim = 80 0 90 0, clip, width =
  0.5\textwidth]{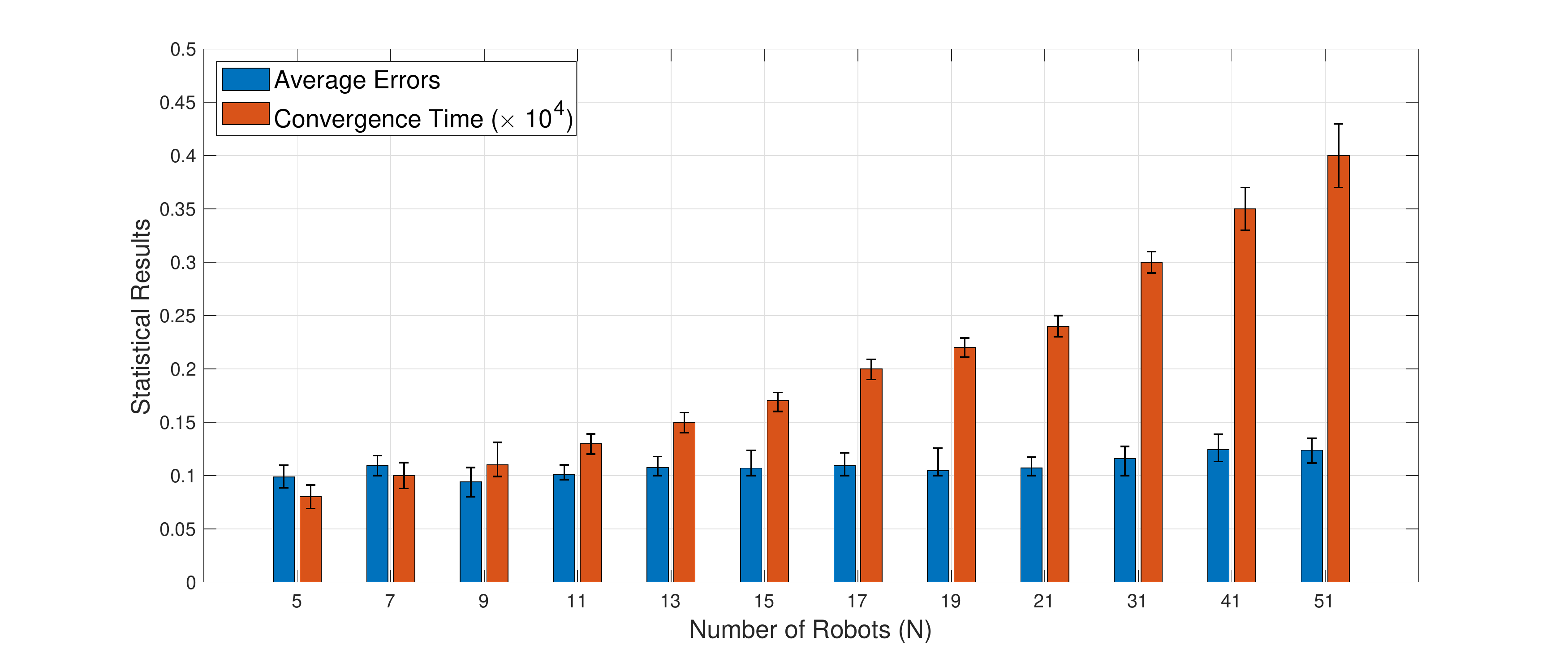}
  \caption{Statistical results for scalability}
  \label{fig:scala}
\end{figure}

\begin{figure}[!htb]
  \centering \includegraphics[trim = 20 200 20 200, clip, width =
  0.4\textwidth]{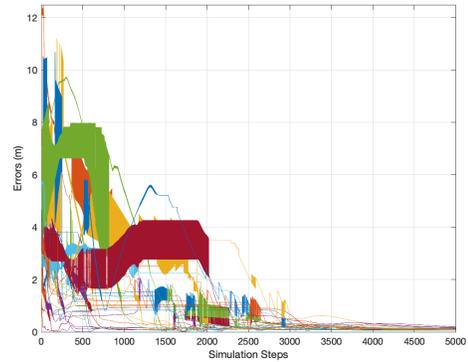}
  \caption{Errors change with 50 followers}
  \label{fig:Error3}
\end{figure}

To test the scalability of the proposed method, we conducted the
simulation under different population sizes of the swarm. The
statistical results are presented in Figure \ref{fig:scala}. We test
the method 10 times for each population configuration, i.e., 5, 7, 9,
11, 13, 15, 17, 19, 21, 31, 41 and 51 respectively. Each bar set in
the figure indicates the average errors after convergence, as well as
the average convergence time. The convergence time is increasing with
the population. However, the final average errors for each swarm is at
the same level. A figure of error change for a 51 robot swarm is shown
in Figure \ref{fig:Error3}. This simulation indicates that the
proposed method is effective under different population size, i.e.
with good scalability.

\section{Conclusion}
This paper proposed an online optimization strategy based on the Brain
Storm Optimization (BSO) algorithm for the coordinated motion control
of swarm robotics. By adopting an incremental PID control law, the
method is able to control the coordinated motion of a swarm of robots
without knowing the leader's velocities. Furthermore, the controller
parameters were optimized online with the adoption of the BSO
algorithm, which can deal with the dynamical target change and
anti-collision requirements. 
The simulation results have demonstrated that the proposed method is
effective for the coordinated motion problem of robotic swarms. The
flexibility and scalability have also been validated to ensure that
the proposed method can adapt to different situations, which makes
this method be a good candidate for coordinated motion control for
swarm robotics under different application scenarios.

\bibliographystyle{IEEEtran} \bibliography{IEEEabrv,mybibfile}

\end{document}